\documentclass[letterpaper, 10 pt, conference]{IEEEconf}
\IEEEoverridecommandlockouts

\usepackage[T1]{fontenc}
\usepackage{amsmath,amsfonts}
\usepackage{array}
\usepackage{cite}
\usepackage[caption=false,font=scriptsize]{subfig}
\usepackage{textcomp}
\usepackage{stfloats}
\usepackage{url}
\usepackage{verbatim}
\usepackage{graphicx}
\usepackage{amssymb}
\usepackage{pifont}
\usepackage{hyperref}
\usepackage[capitalize]{cleveref}
\usepackage{comment}
\usepackage{multirow}
\usepackage{booktabs}
\usepackage{rotating}
\usepackage{makecell}
\usepackage{tabularx}
\usepackage{xcolor}
\usepackage[linesnumbered]{algorithm2e}
\usepackage{setspace}
\RestyleAlgo{ruled}
\DontPrintSemicolon 

\usepackage{siunitx}
\AtBeginDocument{\RenewCommandCopy\qty\SI}
\usepackage{tikz}
\usetikzlibrary{arrows.meta}
\usetikzlibrary{arrows}
\usepackage{pgfplots}
\pgfplotsset{compat=1.8}

\usepackage{xcolor}
\definecolor{Black}{HTML}{000000}
\definecolor{Blue}{HTML}{005293}
\definecolor{Bluestrong}{HTML}{003359}
\definecolor{Red}{HTML}{8C000F}
\definecolor{Orange}{HTML}{E37222}
\definecolor{Green}{HTML}{A2AD00}

\definecolor{visibleArea}{HTML}{9fb99f}
\definecolor{occludedArea}{HTML}{d29f9f}
\definecolor{occludedAreapp}{HTML}{700000}
\definecolor{occludedAreapppropagated}{HTML}{bb4142}

\definecolor{OrangeCR}{HTML}{f1b514}
\definecolor{GreenCR}{HTML}{008000}
\definecolor{GreenForce}{HTML}{00b050}
\definecolor{red}{HTML}{ff0000}
\definecolor{TUMBlue}{HTML}{0065bd}
\definecolor{TUMOrange}{HTML}{E37222}
\definecolor{TUMGreen}{HTML}{9FBA36}
\definecolor{lightOrange}{HTML}{FFC000}
\definecolor{lightGreen}{HTML}{92D050}
\definecolor{GrayArrow}{HTML}{838383}
\definecolor{Grey}{HTML}{808080}
\definecolor{Greylight}{HTML}{CCCCCC}

\newcommand{\visibleArea}{
  \tikz{
    \draw[fill=visibleArea] (0,0.0) rectangle (0.4,0.18);
  }
}

\newcommand{\occludedArea}{
  \tikz{
    \draw[fill=occludedArea] (0,0.0) rectangle (0.4,0.18);
  }
}

\newcommand{\occludedAreapp}{
  \tikz{
    \draw[fill=occludedAreapp] (0,0.0) rectangle (0.4,0.18);
  }
}

\newcommand{\occludedAreapppropagated}{
  \tikz{
    \draw[fill=occludedAreapppropagated] (0,0.0) rectangle (0.4,0.18);
  }
}

\newcommand{\bike}{
  \tikz{
    \node[inner sep=0pt] at (0,0) {\includegraphics[height=3.0mm]{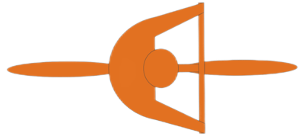}};
  }
}

\newcommand{\phantomBike}{
  \tikz{
    \node[inner sep=0pt] at (0,0) {\includegraphics[height=3.0mm]{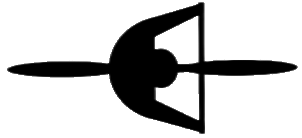}};
  }
}

\newcommand{\phantomPedestrian}{
  \tikz{
    \node[inner sep=0pt] at (0,0) {\includegraphics[height=4.0mm]{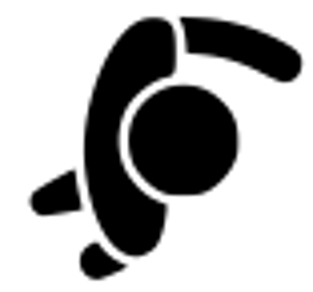}};
  }
}

\newcommand{\phantomPrediction}{
  \tikz{
    \fill[yellow,opacity=0.5] (0,0) circle (0.18);
    \fill[orange,opacity=0.5] (0,0) circle (0.1);
    \fill[red,opacity=0.5] (0,0) circle (0.04);
  }
}

\newcommand{\undetectedBike}{
  \tikz{
    \node[inner sep=0pt] at (0,0) {\includegraphics[height=3.0mm]{input/bike_black.png}};
  }
}

\newcommand{\crashIcon}{
  \tikz{
    \node[inner sep=0pt] at (0,0) {\includegraphics[height=4.8mm]{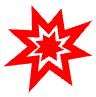}};
  }
}

\newcommand{\crashIconsmall}{
  \tikz{
    \node[inner sep=0pt] at (0,0) {\includegraphics[height=3.8mm]{input/crashIcon.png}};
  }
}

\def\BibTeX{{\rm B\kern-.05em{\sc i\kern-.025em b}\kern-.08em
    T\kern-.1667em\lower.7ex\hbox{E}\kern-.125emX}}

\begin{document}


\author{Korbinian Moller, Luis Schwarzmeier, Johannes Betz%
\thanks{K. Moller, L. Schwarzmeier and J. Betz are with the Professorship of Autonomous Vehicle Systems, TUM School of Engineering and Design, Technical University of Munich, 85748 Garching, Germany; Munich Institute of Robotics and Machine Intelligence (MIRMI).}%
}
%
%
%



\title{\LARGE \bf From Shadows to Safety: Occlusion Tracking and Risk Mitigation for Urban Autonomous Driving}
\maketitle


\begin{abstract}

Autonomous vehicles (AVs) must navigate dynamic urban environments where occlusions and perception limitations introduce significant uncertainties. This research builds upon and extends existing approaches in risk-aware motion planning and occlusion tracking to address these challenges. While prior studies have developed individual methods for occlusion tracking and risk assessment, a comprehensive method integrating these techniques has not been fully explored. We, therefore, enhance a phantom agent-centric model by incorporating sequential reasoning to track occluded areas and predict potential hazards. Our model enables realistic scenario representation and context-aware risk evaluation by modeling diverse phantom agents, each with distinct behavior profiles. Simulations demonstrate that the proposed approach improves situational awareness and balances proactive safety with efficient traffic flow. While these results underline the potential of our method, validation in real-world scenarios is necessary to confirm its feasibility and generalizability. By utilizing and advancing established methodologies, this work contributes to safer and more reliable AV planning in complex urban environments. To support further research, our method is available as open-source software at \url{https://github.com/TUM-AVS/OcclusionAwareMotionPlanning}.

\end{abstract}


\vspace{0.1cm}
\begin{keywords}
Autonomous Driving, Motion Planning, Safety, Occlusion Awareness, Vulnerable Road User
\end{keywords}
%


\section{Introduction}
\label{sec:introduction}

Autonomous vehicles (AVs) hold the potential to revolutionize transportation by reducing accidents, enhancing mobility, and improving traffic efficiency~\cite{Herrmann2018}. However, their safe deployment in dynamic urban environments remains a considerable challenge, particularly when navigating scenarios involving vulnerable road users (VRUs) such as pedestrians and cyclists. Despite significant advancements, perception systems in AVs are inherently limited by constraints such as sensor range, field of view, and occlusions caused by static obstacles (e.g., parked vehicles or buildings) and dynamic objects (e.g., moving vehicles)~\cite{Yang2023}. These limitations result in hidden areas within the environment, introducing uncertainties that have been shown to contribute significantly to crashes~\cite{Singh2018}.

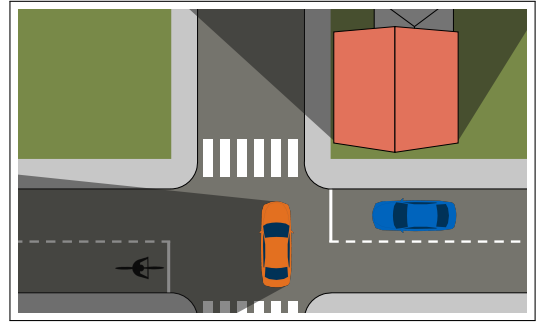
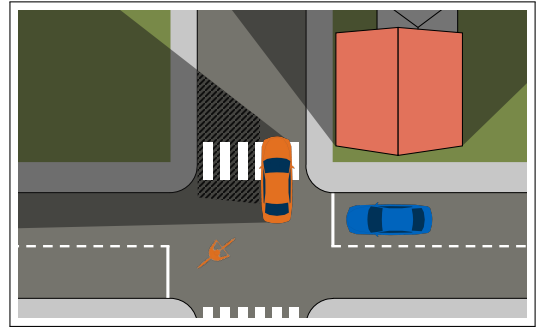
\begin{figure}[!t]
    \centering
    \hspace{1mm}
    \begin{tikzpicture}[font=\scriptsize]
        \node[inner sep=0pt] at (0.9,0) {\includegraphics[height=2.5mm]{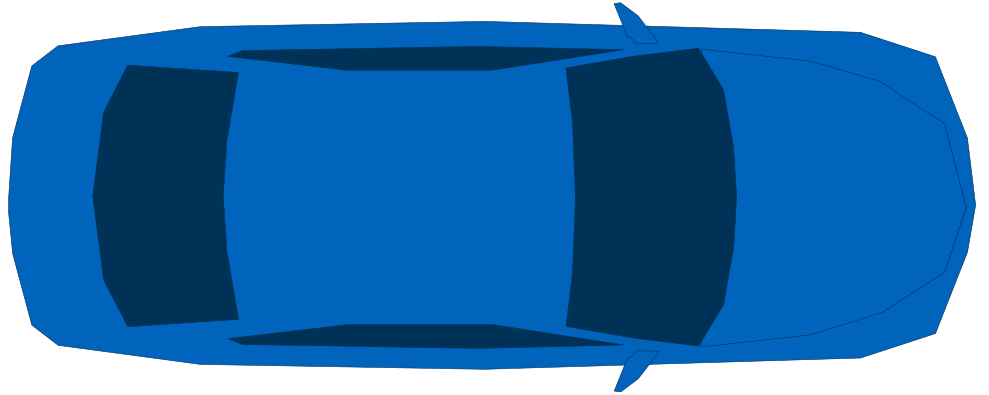}};
        \node[align=left, anchor=west] at (1.2,0) {ego \\ vehicle};
    
        \node[inner sep=0pt] at (2.7,0) {\includegraphics[height=2.5mm]{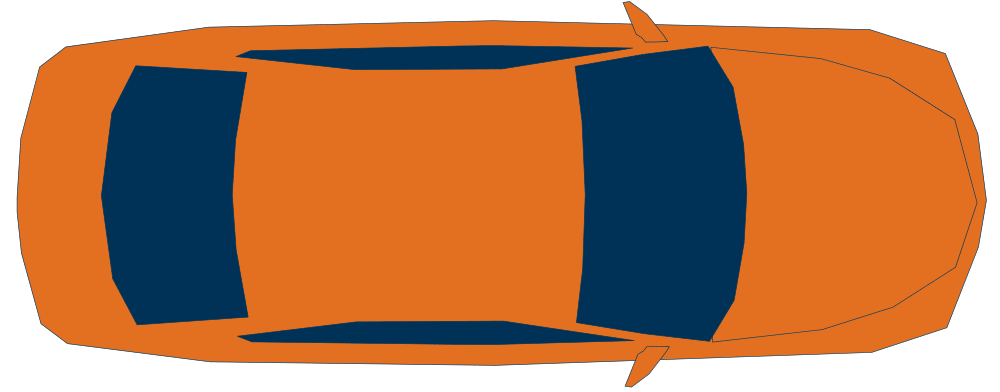}};
        \node[align=left, anchor=west] at (3.0,0) {dynamic \\ obstacle};

        \node[inner sep=0pt] at (4.5,0) {\undetectedBike};
        \node[align=left, anchor=west] at (4.7,0) {undetected \\ cyclist};

        \node[inner sep=0pt] at (6.4,0) {\bike};
        \node[align=left, anchor=west] at (6.6,0) {detected \\ cyclist};

        
    \end{tikzpicture}
    \subfloat[][The cyclist is located in an occluded area and is not yet visible to the ego vehicle (EV).]{\label{fig:introduction_1} \fbox{\includegraphics[width=0.38\textwidth, trim={5cm 13cm 3cm 7cm},clip]{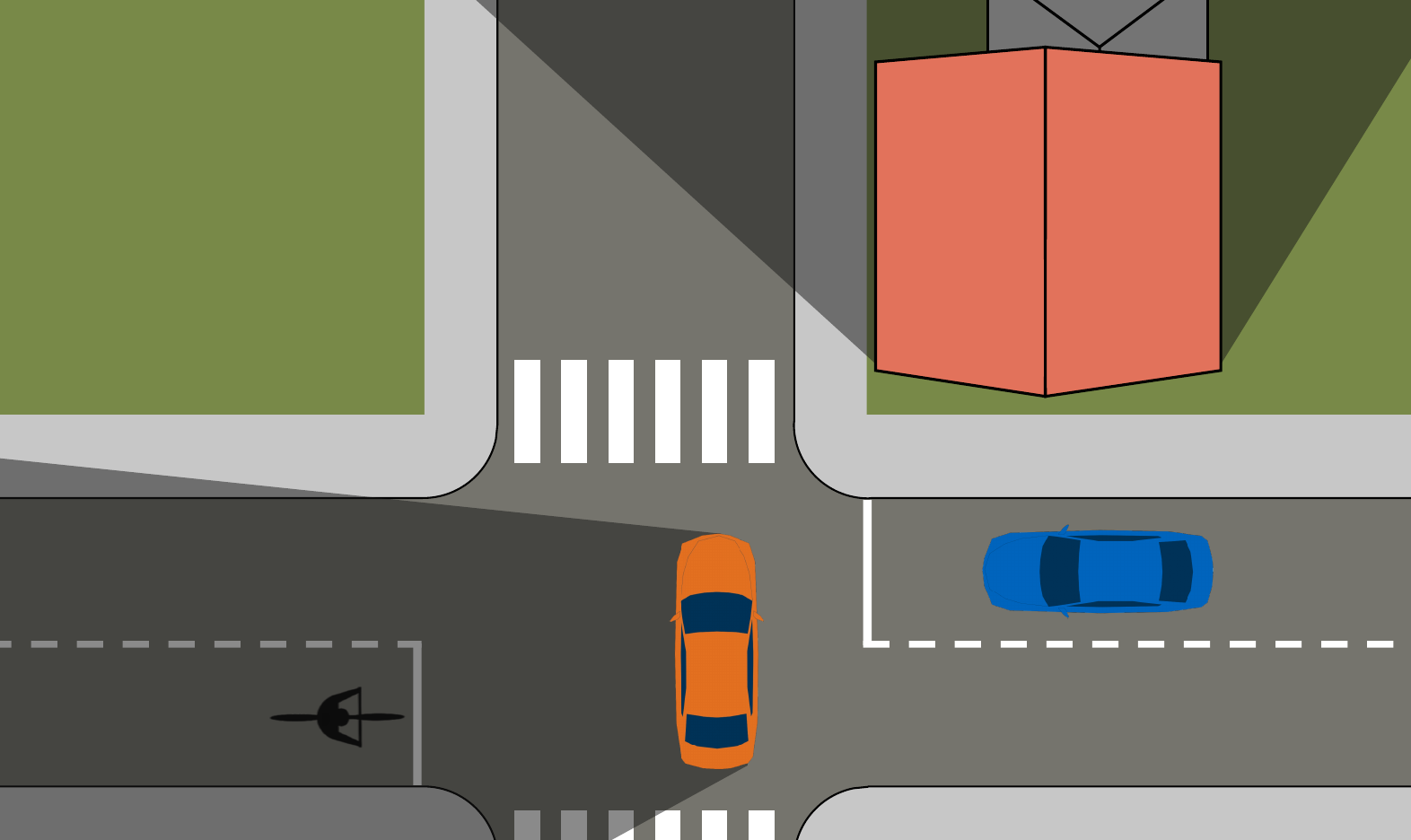}}} \\
    \subfloat[][The cyclist emerges from the occluded area, becoming visible to the EV. The hatched region indicates an area where no objects can be present, as it was previously visible to the EV.]{\label{fig:introduction_2} \fbox{\includegraphics[width=0.38\textwidth, trim={5cm 13cm 3cm 7cm},clip]{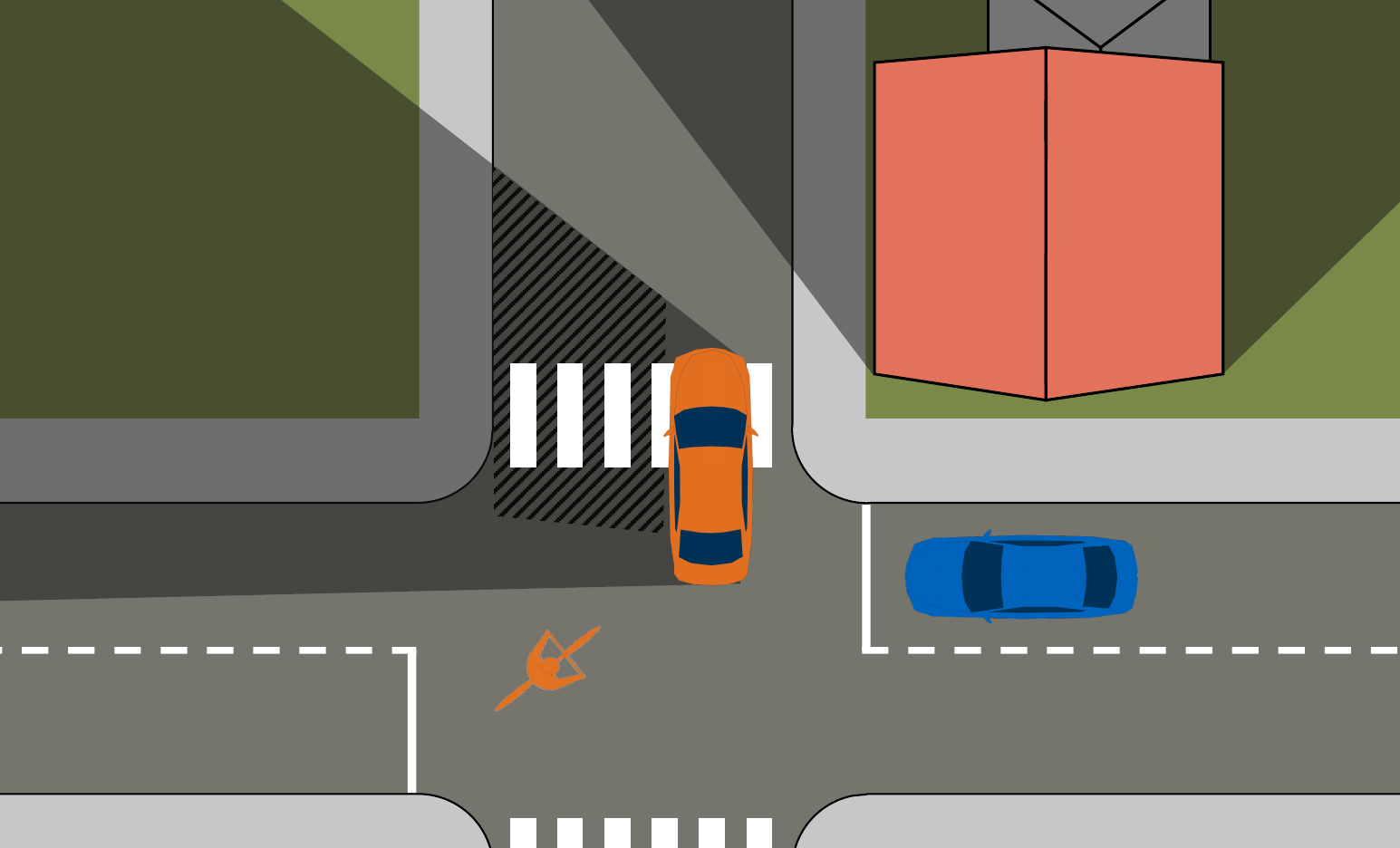}}} \\
    \caption{Illustration of an intersection scenario emphasizing the importance of occlusion-aware planning.}
    \label{fig:introduction}%
\end{figure}

To ensure safe navigation despite these perception constraints, AVs must incorporate methods that compensate for limited perception capabilities. This involves dynamically assessing occluded areas and identifying potential hazards to enhance situational awareness and mitigate risks associated with occluded road users. By doing so, AVs can effectively balance proactive safety measures with efficient traffic flow. Figure~\ref{fig:introduction} illustrates a typical scenario where a cyclist emerges from an occluded area, underscoring the critical importance of occlusion-aware planning.

This paper addresses these challenges by proposing a motion planning algorithm that integrates and extends existing risk- and occlusion-aware motion planning approaches. Central to this approach is the enhancement of a phantom agent-centric model~\cite{Moller2024} to systematically track occluded areas to identify potential spawn points for occluded traffic participants. Sequential reasoning~\cite{Wang2021-reasoning, Sanchez2022} is employed as part of this tracking process, dynamically assessing which portions of the occluded area could realistically be reached by hidden agents based on their potential paths and speeds. This integration ensures that only relevant phantom agents (PAs) are considered in the planning process, reducing unnecessary computational overhead while focusing on real risks~\cite{Moller2024}. The proposed method introduces diverse PA types, representing vehicles, pedestrians, and cyclists, each characterized by distinct behavior profiles. This diversity enables the planner to account for a wide range of potential actions and interactions. In conclusion, our proposed occlusion-aware motion planner presents four main contributions:

\begin{enumerate}
    \item \textbf{A combination of agent-centric modeling and sequential reasoning:} This work integrates occlusion tracking with sequential reasoning to dynamically evaluate reachable areas in occluded zones. By focusing on PAs in relevant areas, only those traffic participants that pose a realistic risk are considered.
    \item \textbf{A diverse PA framework for behavior modeling:} Our method introduces different PA types with various behavior profiles. This enables the representation of a wide range of possible actions, capturing diverse and realistic agent behaviors.
    \item \textbf{Comprehensive simulation analysis:} Extensive simulations are conducted to evaluate the proposed approach in real-world-inspired scenarios. These analyses demonstrate the effectiveness of our method in improving occlusion awareness.
    \item \textbf{Open-source software:} The occlusion-aware motion planner, including occlusion tracking methods and PA models, is published as open-source software.
\end{enumerate}

\section{Related Work}
\label{sec:relatedwork}

The challenge of navigating environments with occluded obstacles has prompted a variety of approaches in autonomous driving research, each addressing different aspects of the problem.

One class of solutions employs Partially Observable Markov Decision Processes (POMDPs) to model the uncertainties associated with occluded areas. These methods reason about hidden traffic participants based on their potential trajectories and interactions~\cite{Zhang2022, Wray2021}. While POMDP-based approaches provide robust theoretical frameworks, their practical application often involves high computational complexity, limiting real-time feasibility.

Another strategy involves reachable sets, which predict the possible positions and velocities of occluded road users based on formalized traffic rules and motion models~\cite{Koschi2021, Orzechowski2018}. These approaches have demonstrated versatility across various traffic scenarios, including urban intersections and autonomous parking maneuvers~\cite{Lee2021}.

Several studies aim to mitigate occlusion risks by increasing the visible area. For instance, lateral position adjustments can expand sensor coverage, enabling AVs to detect otherwise occluded road users. Cost functions that prioritize visibility in trajectory planning foster collision avoidance~\cite{Gilhuly2022, Narksri2022}. Similarly, infrastructure-based enhancements, such as roadside units (RSUs), extend the vehicle’s perception range by integrating local and external sensors~\cite{Borba2025}. In~\cite{Nyberg2024} the authors leveraged Vehicle-to-Everything (V2X) technology to improve visibility through sequential reasoning, demonstrating potential for urban environments. Despite these advancements, such methods depend on the availability of external infrastructure, which is often not feasible in current real-world scenarios.

In~\cite{Trentin2024}, the authors propose a Dynamic Bayesian Network with Markov chains to tackle occlusions. By generating multimodal predictions and evaluating these in realistic use cases, the framework advances occlusion-aware planning.

Previous research has also explored methods to address agents that temporarily disappear. Pang et al.~\cite{Pang2023} introduced a motion prediction framework that propagates past positions of occluded agents using multi-modal trajectory predictions and differentiable filters to ensure temporal coherence.

Another line of research models occluded road users as PAs. Zhao et al.~\cite{Zhao2024} introduced a framework that uses PAs to infer vehicle trajectories at intersections by comparing predicted and concatenated paths. However, this approach does not account for pedestrians or cyclists, whose behavior patterns differ significantly. Zhong et al.~\cite{Zhong2023} extended this concept by incorporating probabilistic models that estimate PA existence based on occlusion duration and proximity to the ego vehicle (EV).


Sequential reasoning further enhances occlusion-aware planning by dynamically eliminating unrealistic obstacle states~\cite{Wang2021-reasoning, Sanchez2022}. This methodology integrates reachable set analysis with motion prediction. Risk-aware motion planning, on the other hand, focuses on estimating potential dangers posed by occluded traffic participants. This approach is particularly relevant in urban scenarios involving pedestrians, aiming to minimize collisions or potential harm~\cite{Wang2023, Koc2021, TrauthOJITS}.

Building on these concepts, our earlier work~\cite{Moller2024} proposed an algorithm that integrates PA modeling with criticality metrics to assess potential risks from occluded areas. By dynamically evaluating trajectory safety and incorporating modular integration into motion planning algorithms, this approach demonstrated significant improvements in balancing proactive safety and traffic efficiency.

While the presented approaches address individual aspects of occlusion-aware planning, their isolated application is insufficient for comprehensive safety in urban environments. Reachability-based methods lack the ability to distinguish between different traffic participants, such as pedestrians, cyclists, and vehicles, in terms of their motion dynamics and associated collision risks. Conversely, agent-centric approaches often overlook the geometric feasibility of occluded areas.

\section{Methodology}
\label{sec:method}

The limitations of existing approaches, as outlined, highlight the need for an integrated solution that combines the strengths of reachability-based and agent-centric methods. We, therefore, extend the open-source framework from our earlier work~\cite{Moller2024}. Our new approach enhances the algorithm by dynamically tracking visible areas~\cite{Wang2021-reasoning, Sanchez2022} to ensure PAs are placed only in critical occluded areas, extending predictions for previously visible objects, and integrating risk assessments tailored to diverse road user dynamics.


\subsection{Overview of the Proposed Methodology}

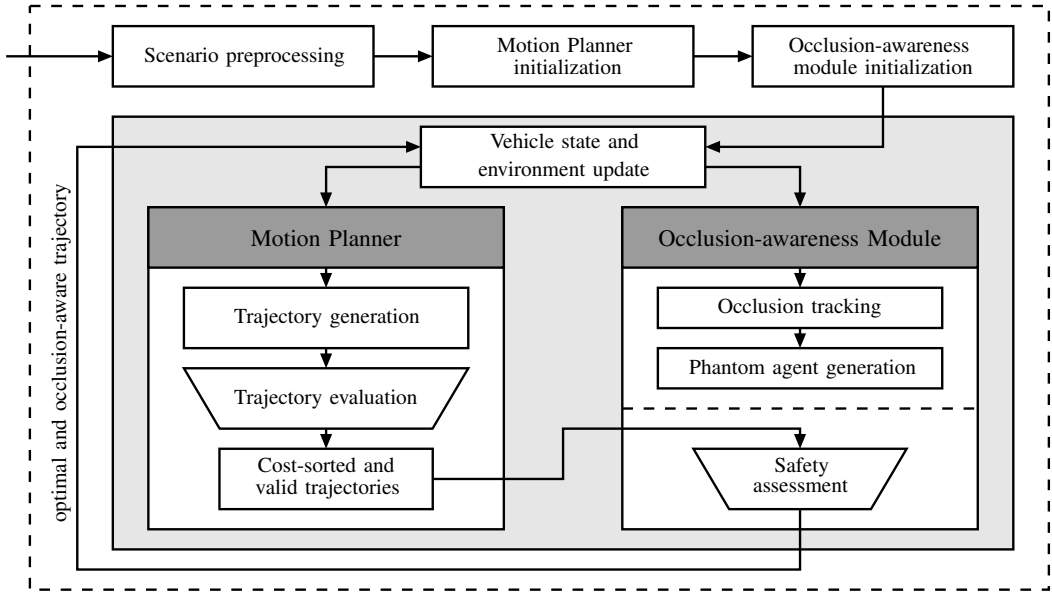
\begin{figure*}[!ht]
    \centering
    %
    \resizebox{0.80\textwidth}{!}{

\tikzset{every picture/.style={line width=1.0pt}} 

\begin{tikzpicture}[x=1.0pt,y=0.85pt,yscale=-1,xscale=1]

\draw  [color={rgb, 255:red, 0; green, 0; blue, 0 }  ,draw opacity=1 ][fill={rgb, 255:red, 230; green, 230; blue, 230 }  ,fill opacity=1 ] (145,135) -- (525,135) -- (525,350) -- (145,350) -- cycle ;
\draw  [color={rgb, 255:red, 0; green, 0; blue, 0 }  ,draw opacity=1 ][dash pattern={on 4.5pt off 4.5pt}] (110,80) -- (540,80) -- (540,370) -- (110,370) -- cycle ;
\draw    (100,105) -- (142,105) ;
\draw [shift={(145,105)}, rotate = 180] [fill={rgb, 255:red, 0; green, 0; blue, 0 }  ][line width=0.08]  [draw opacity=0] (6.25,-3) -- (0,0) -- (6.25,3) -- cycle    ;
\draw   (145,90) -- (255,90) -- (255,120) -- (145,120) -- cycle ;
\draw  [color={rgb, 255:red, 0; green, 0; blue, 0 }  ,draw opacity=1 ][fill={rgb, 255:red, 255; green, 255; blue, 255 }  ,fill opacity=1 ] (160,180) -- (310,180) -- (310,340) -- (160,340) -- cycle ;
\draw   (280,90) -- (390,90) -- (390,120) -- (280,120) -- cycle ;
\draw    (255,105) -- (277,105) ;
\draw [shift={(280,105)}, rotate = 180] [fill={rgb, 255:red, 0; green, 0; blue, 0 }  ][line width=0.08]  [draw opacity=0] (6.25,-3) -- (0,0) -- (6.25,3) -- cycle    ;
\draw   (415,90) -- (525,90) -- (525,120) -- (415,120) -- cycle ;
\draw    (390,105) -- (412,105) ;
\draw [shift={(415,105)}, rotate = 180] [fill={rgb, 255:red, 0; green, 0; blue, 0 }  ][line width=0.08]  [draw opacity=0] (6.25,-3) -- (0,0) -- (6.25,3) -- cycle    ;
\draw  [color={rgb, 255:red, 0; green, 0; blue, 0 }  ,draw opacity=1 ][fill={rgb, 255:red, 255; green, 255; blue, 255 }  ,fill opacity=1 ] (360,180) -- (510,180) -- (510,340) -- (360,340) -- cycle ;
\draw  [fill={rgb, 255:red, 255; green, 255; blue, 255 }  ,fill opacity=1 ] (275,140) -- (395,140) -- (395,170) -- (275,170) -- cycle ;
\draw    (470,120) -- (470,150) -- (398,150) ;
\draw [shift={(395,150)}, rotate = 360] [fill={rgb, 255:red, 0; green, 0; blue, 0 }  ][line width=0.08]  [draw opacity=0] (6.25,-3) -- (0,0) -- (6.25,3) -- cycle    ;
\draw   (175,220) -- (295,220) -- (295,250) -- (175,250) -- cycle ;
\draw    (275,160) -- (235,160) -- (235,177) ;
\draw [shift={(235,180)}, rotate = 270] [fill={rgb, 255:red, 0; green, 0; blue, 0 }  ][line width=0.08]  [draw opacity=0] (6.25,-3) -- (0,0) -- (6.25,3) -- cycle    ;
\draw    (395,160) -- (435,160) -- (435,177) ;
\draw [shift={(435,180)}, rotate = 270] [fill={rgb, 255:red, 0; green, 0; blue, 0 }  ][line width=0.08]  [draw opacity=0] (6.25,-3) -- (0,0) -- (6.25,3) -- cycle    ;
\draw  [color={rgb, 255:red, 0; green, 0; blue, 0 }  ,draw opacity=1 ][fill={rgb, 255:red, 155; green, 155; blue, 155 }  ,fill opacity=1 ] (160,180) -- (310,180) -- (310,210) -- (160,210) -- cycle ;
\draw  [color={rgb, 255:red, 0; green, 0; blue, 0 }  ,draw opacity=1 ][fill={rgb, 255:red, 155; green, 155; blue, 155 }  ,fill opacity=1 ] (360,180) -- (510,180) -- (510,210) -- (360,210) -- cycle ;
\draw    (235,210) -- (235,217) ;
\draw [shift={(235,220)}, rotate = 270] [fill={rgb, 255:red, 0; green, 0; blue, 0 }  ][line width=0.08]  [draw opacity=0] (6.25,-3) -- (0,0) -- (6.25,3) -- cycle    ;
\draw    (235,250) -- (235,257) ;
\draw [shift={(235,260)}, rotate = 270] [fill={rgb, 255:red, 0; green, 0; blue, 0 }  ][line width=0.08]  [draw opacity=0] (6.25,-3) -- (0,0) -- (6.25,3) -- cycle    ;
\draw   (375,220) -- (495,220) -- (495,240) -- (375,240) -- cycle ;
\draw   (375,250) -- (495,250) -- (495,270) -- (375,270) -- cycle ;
\draw    (435,210) -- (435,217) ;
\draw [shift={(435,220)}, rotate = 270] [fill={rgb, 255:red, 0; green, 0; blue, 0 }  ][line width=0.08]  [draw opacity=0] (6.25,-3) -- (0,0) -- (6.25,3) -- cycle    ;
\draw    (435,240) -- (435,247) ;
\draw [shift={(435,250)}, rotate = 270] [fill={rgb, 255:red, 0; green, 0; blue, 0 }  ][line width=0.08]  [draw opacity=0] (6.25,-3) -- (0,0) -- (6.25,3) -- cycle    ;
\draw    (435,330) -- (435,360) -- (130,360) -- (130,150) -- (272,150) ;
\draw [shift={(275,150)}, rotate = 180] [fill={rgb, 255:red, 0; green, 0; blue, 0 }  ][line width=0.08]  [draw opacity=0] (6.25,-3) -- (0,0) -- (6.25,3) -- cycle    ;
\draw   (175,260) -- (190.08,290) -- (279.92,290) -- (295,260) -- cycle ;
\draw   (190,300) -- (280,300) -- (280,330) -- (190,330) -- cycle ;
\draw    (235,290) -- (235,297) ;
\draw [shift={(235,300)}, rotate = 270] [fill={rgb, 255:red, 0; green, 0; blue, 0 }  ][line width=0.08]  [draw opacity=0] (6.25,-3) -- (0,0) -- (6.25,3) -- cycle    ;
\draw  [dash pattern={on 4.5pt off 4.5pt}]  (360,280) -- (510,280) ;
\draw   (390,300) -- (405.08,330) -- (464.92,330) -- (480,300) -- cycle ;
\draw    (280,315) -- (335,315) -- (335,290) -- (435,290) -- (435,297) ;
\draw [shift={(435,300)}, rotate = 270] [fill={rgb, 255:red, 0; green, 0; blue, 0 }  ][line width=0.08]  [draw opacity=0] (6.25,-3) -- (0,0) -- (6.25,3) -- cycle    ;

\draw (157,99) node [anchor=north west][inner sep=0.75pt]   [align=left] {{\small Scenario preprocessing}};
\draw (306,94) node [anchor=north west][inner sep=0.75pt]   [align=left] {{\small Motion Planner}};
\draw (313,105) node [anchor=north west][inner sep=0.75pt]   [align=left] {{\small initialization}};
\draw (429,94) node [anchor=north west][inner sep=0.75pt]   [align=left] {{\small Occlusion-awareness}};
\draw (431,105) node [anchor=north west][inner sep=0.75pt]   [align=left] {{\small module initialization}};
\draw (303,142) node [anchor=north west][inner sep=0.75pt]   [align=left] {{\small Vehicle state and}};
\draw (195,229) node [anchor=north west][inner sep=0.75pt]   [align=left] {{\small Trajectory generation}};
\draw (202,190) node [anchor=north west][inner sep=0.75pt] [align=left] {Motion Planner};
\draw (374,190) node [anchor=north west][inner sep=0.75pt] [align=left] {Occlusion-awareness Module};
\draw (195,269) node [anchor=north west][inner sep=0.75pt]   [align=left] {{\small Trajectory evaluation}};
\draw (399,224) node [anchor=north west][inner sep=0.75pt]   [align=left] {{\small Occlusion tracking}};
\draw (387,254) node [anchor=north west][inner sep=0.75pt]   [align=left] {{\small Phantom agent generation}};
\draw (118,338) node [anchor=north west][inner sep=0.75pt]  [rotate=-270.0] [align=left] {{\small optimal and occlusion-aware trajectory}};
\draw (205,303) node [anchor=north west][inner sep=0.75pt]   [align=left] {{\small Cost-sorted and }};
\draw (298,156) node [anchor=north west][inner sep=0.75pt]   [align=left] {{\small environment update}};
\draw (204,314) node [anchor=north west][inner sep=0.75pt]   [align=left] {{\small valid trajectories }};
\draw (423,303) node [anchor=north west][inner sep=0.75pt]   [align=left] {{\small Safety}};
\draw (414,314) node [anchor=north west][inner sep=0.75pt]   [align=left] {{\small assessment}};

\end{tikzpicture}
}%

    \caption{Overview of the framework integrating a motion planner with our occlusion-awareness module. The motion planner generates and evaluates trajectories, providing cost-sorted and valid trajectories. The occlusion-awareness module identifies potential risks in occluded areas $\mathcal{A}_\mathrm{o}^\mathcal{L}$ and performs a safety assessment. This ensures that an optimal and occlusion-aware trajectory is selected.}
    \label{fig:framework}
\end{figure*}

The proposed algorithm is designed as a modular evaluation layer that operates with existing motion planning algorithms, as shown in \Cref{fig:framework}. It evaluates candidate trajectories generated by the planner with respect to safety in occluded areas. To achieve this, specific information about the environment, the EV, and surrounding traffic participants is required.

The trajectories to be evaluated must be provided in global $(x, y)$ coordinates as a time-parameterized sequence:
\begin{equation}
    \xi(t) = \{(x_1, y_1, v_1)^{\xi}, (x_2, y_2, v_2)^{\xi}, \dots, (x_n, y_n, v_n)^{\xi}\},
\end{equation}
where $v_i$ represents the velocity at each position $(x_i, y_i)$. Additionally, a global reference path $\Gamma$ is required to assess the relevance of occluded areas along the planned route:
\begin{equation}
    \Gamma = \{(x_1, y_1)^{\Gamma}, (x_2, y_2)^{\Gamma}, \dots, (x_m, y_m)^{\Gamma}\}.
\end{equation}

The state of the EV and other traffic participants is defined by their global position $(x, y)$, orientation $\theta$ and velocity $v$ as well as the progress $s$ along and the lateral deviation $d$ from $\Gamma$:
\begin{equation}
    \mathbf{X} = 
    \begin{bmatrix}
        x, y, \theta, v, s, d
    \end{bmatrix}^\intercal.
\end{equation}

The road network $\mathcal{L}$ is modeled as a set of lanelets $\ell$ ~\cite{Bender2014}, where each lanelet $\ell_i \in \mathcal{L}$ is defined by its left boundary $\mathbf{b}_\mathrm{l}$, right boundary $\mathbf{b}_\mathrm{r}$, and a set of constraints $\mathcal{C}^\ell$:
\begin{equation}
    \ell_i = \{ \mathbf{b}_\mathrm{l}, \mathbf{b}_\mathrm{r}, \mathcal{C}^\ell \}.
\end{equation}

The boundaries $\mathbf{b}_\mathrm{l}$ and $\mathbf{b}_\mathrm{r}$ are represented as sequences of points:
\begin{equation}
    \mathbf{b} = \{(x_1, y_1)^{\mathbf{b}}, (x_2, y_2)^{\mathbf{b}}, \dots\}.
\end{equation}

Each lanelet $\ell_i$ includes specific constraints $\mathcal{C}^\ell$, such as the road type (\textit{e.g.}, urban, highway) and maximum permissible velocity $\text{v}_{\mathrm{max}}$. To define valid positions within a lanelet, we introduce the set $\mathcal{X}^\ell \subset \mathbb{R}^{2} $, which contains all points that belong to the lanelet $\ell$. From $\mathcal{X}^\ell$, we define the set of valid states $\mathcal{X}_v^\ell$, which respect all constraints $\mathcal{C}^\ell$ associated with the lanelet:
\begin{equation}
    \mathcal{X}_v^\ell = \{x \in \mathcal{X}^\ell \,|\, x \in \mathcal{C}^\ell \}.
\end{equation}

These valid states $\mathcal{X}_v^\ell$ are critical for applications such as tracking occluded areas, as they ensure that the movement within the lanelet complies with the defined constraints, including the maximum permissible velocity $\text{v}_{\mathrm{max}}$.

By combining the EV’s state, the positions of surrounding traffic participants, and the structured road network, the framework dynamically identifies occluded areas, evaluates their relevance to the planned trajectory, and assesses the safety of candidate motion plans.


\subsection{Visible Area Calculation and Occlusion Tracking}

The visible area $\mathcal{A}_\mathrm{v}^{\mathcal{L}} \subset \mathbb{R}^{2}$ is required for identifying and updating occluded areas in the environment. It is computed at every timestep $t (\Delta t = \SI{0.1}{\second})$ based on the EV's state $\mathbf{X}_\mathrm{EV}$ and the surrounding environment, including obstacles $\mathcal{O}$ and the road network $\mathcal{L}$, as shown in Algorithm~\ref{algo:VisibleAreaCalc}.
\begin{algorithm}[!ht]
\SetAlgoLined
\setstretch{1.1}
\SetKwInOut{Input}{Input}\SetKwInOut{Output}{Output}
\SetKwFunction{CalcRoadArea}{computeRoadArea}
\SetKwFunction{CalcObstacleShadows}{subtractObstacleShadows}
\SetKwFunction{UpdateObstacleState}{updateObstacleState}

\Input{Ego Vehicle State $\mathbf{X}_\mathrm{EV}$, Obstacles $\mathcal{O}$, Road Network $\mathcal{L}$, Sensor Range $\mathcal{A}_\mathrm{r}$}
\Output{Visible Area $\mathcal{A}_\mathrm{v}^\mathcal{L}$, Visible Obstacles $\mathcal{O}_\mathrm{v}$}


$\mathcal{A}_\mathrm{v}^\mathcal{L} \leftarrow$ \CalcRoadArea{$\mathbf{X}_\mathrm{EV}, \mathcal{A}_\mathrm{r}, \mathcal{L}$}\;

$\mathcal{A}_\mathrm{v}^\mathcal{L} \leftarrow$ \CalcObstacleShadows{$\mathcal{A}_\mathrm{v}^\mathcal{L}, \mathcal{O}$}\;

$\mathcal{O}_\mathrm{v} \leftarrow \{\}$\;

\ForEach{$o \in \mathcal{O}$}{
    \If{$o.\mathrm{polygon} \cap \mathcal{A}_\mathrm{v}^\mathcal{L} \neq \emptyset$}{
        $\mathcal{O}_\mathrm{v} \leftarrow \mathcal{O}_\mathrm{v} \cup \{o\}$\;
    }
}

\Return $\mathcal{A}_\mathrm{v}^\mathcal{L}, \mathcal{O}_\mathrm{v}$\;

\caption{Visible Area and Obstacle Detection \label{algo:VisibleAreaCalc}}
\end{algorithm}

$\mathcal{A}_\mathrm{v}^{\mathcal{L}}$ is defined as the region within the sensor's range $\mathcal{A}_\mathrm{r}$, restricted to the area of the road network $\mathcal{A}_\mathcal{L}$, excluding regions occluded by obstacles:
\begin{equation}
    \mathcal{A}_\mathrm{v}^{\mathcal{L}} = \big( \mathcal{A}_\mathrm{r} \cap \mathcal{A}_\mathcal{L} \big) \setminus \bigcup_{o \in \mathcal{O}} \mathcal{A}_\mathrm{o}^{o},
\end{equation}
where $\mathcal{A}_\mathrm{o}^{o}$ represents the shadow area of an obstacle $o \in \mathcal{O}$, determined by the obstacle's geometry and its relative position to the EV. 

Based on $\mathcal{A}_\mathrm{v}^\mathcal{L}$, sequential reasoning is deployed to track and update occluded areas $\mathcal{A}_\mathrm{o}^\mathcal{L}$ in the environment. The process is outlined in Algorithm~\ref{algo:UpdateTracker}.
\begin{algorithm}[!ht]
\SetAlgoLined
\setstretch{1.1}
\SetKwInOut{Input}{Input}\SetKwInOut{Output}{Output}
\SetKwFunction{InitializeLaneOcclusions}{initOcclusions}
\SetKwFunction{ExpandLaneOcclusion}{expandOcclusion}

\Input{Visible Area $\mathcal{A}_\mathrm{v}^\mathcal{L}$, Road Network $\mathcal{L}$}
\Output{Updated Occluded Area $\mathcal{A}_\mathrm{o}^\mathcal{L}$}


\ForEach{$\ell \in \mathcal{L}$}{
    \uIf{$\mathcal{A}_\mathrm{o}^\ell = \emptyset$}{
        $\mathcal{A}_\mathrm{o}^\ell \leftarrow$ \InitializeLaneOcclusions{$\mathcal{A}_\mathrm{v}^\ell, \mathcal{X}_v^\ell$}\;
    }
    \Else{
        $\mathcal{A}_\mathrm{o}^\ell \leftarrow$ \ExpandLaneOcclusion{$\mathcal{A}_\mathrm{o}^\ell, \mathcal{X}_v^\ell, \mathcal{C}^\ell$}\;
        $\mathcal{A}_\mathrm{o}^\ell \leftarrow \mathcal{A}_\mathrm{o}^\ell \setminus \mathcal{A}_\mathrm{v}^\ell$\;
    }
}

\Return $\mathcal{A}_\mathrm{o}^\mathcal{L} \leftarrow \bigcup_{\ell \in \mathcal{L}} \mathcal{A}_\mathrm{o}^\ell$\;

\caption{Occlusion Tracking and Update \label{algo:UpdateTracker}}
\end{algorithm}

At the initial timestep $t_\mathrm{0} = 0$, when no prior observations are available, the occluded area $\mathcal{A}_\mathrm{o}^\ell$ of a lane $\ell$ is determined by the valid states $\mathcal{X}_v^\ell$ of the lane that lie outside the visible area $\mathcal{A}_\mathrm{v}^\ell$:
\begin{equation}
    \mathcal{A}_\mathrm{o}^\ell = \{ x \in \mathcal{X}_v^\ell \;|\; x \notin \mathcal{A}_\mathrm{v}^\ell \}.
\end{equation}
This initialization captures all areas within the lane $\ell$ that are not visible at $t_\mathrm{0}$, and the total occluded area is given by the union of all initial lane occlusions:
\begin{equation}
    \mathcal{A}_\mathrm{o}^\mathcal{L} = \bigcup_{\ell \in \mathcal{L}} \mathcal{A}_\mathrm{o}^\ell.
    \label{eq:A_o_L}
\end{equation}

\begin{figure}[!t]
    \centering
    \hspace{1mm}
    \begin{tikzpicture}[font=\scriptsize]
        \node[inner sep=0pt] at (0.9,0) {\includegraphics[height=2.5mm]{input/ego.png}};
        \node[align=left, anchor=west] at (1.2,0) {ego \\ vehicle};
    
        \node[inner sep=0pt] at (2.7,0) {\includegraphics[height=2.5mm]{input/dyn_obstacle.png}};
        \node[align=left, anchor=west] at (3.0,0) {dynamic \\ obstacle};

        \node[inner sep=0pt] at (4.5,0) {\occludedAreapp};
        \node[align=left, anchor=west] at (4.7,0) {$\mathcal{A}_\mathrm{o}^\mathcal{L}$};

        \node[inner sep=0pt] at (5.7,0) {\occludedAreapppropagated};
        \node[align=left, anchor=west] at (5.9,0) {propagated $\mathcal{A}_\mathrm{o}^\mathcal{L}$};
    \end{tikzpicture}
    \subfloat[][Scenario at the initial timestep without prior sensor updates.]{\label{fig:tracking_0} \fbox{\includegraphics[width=0.38\textwidth, trim={2cm 13cm 3cm 8.5cm},clip]{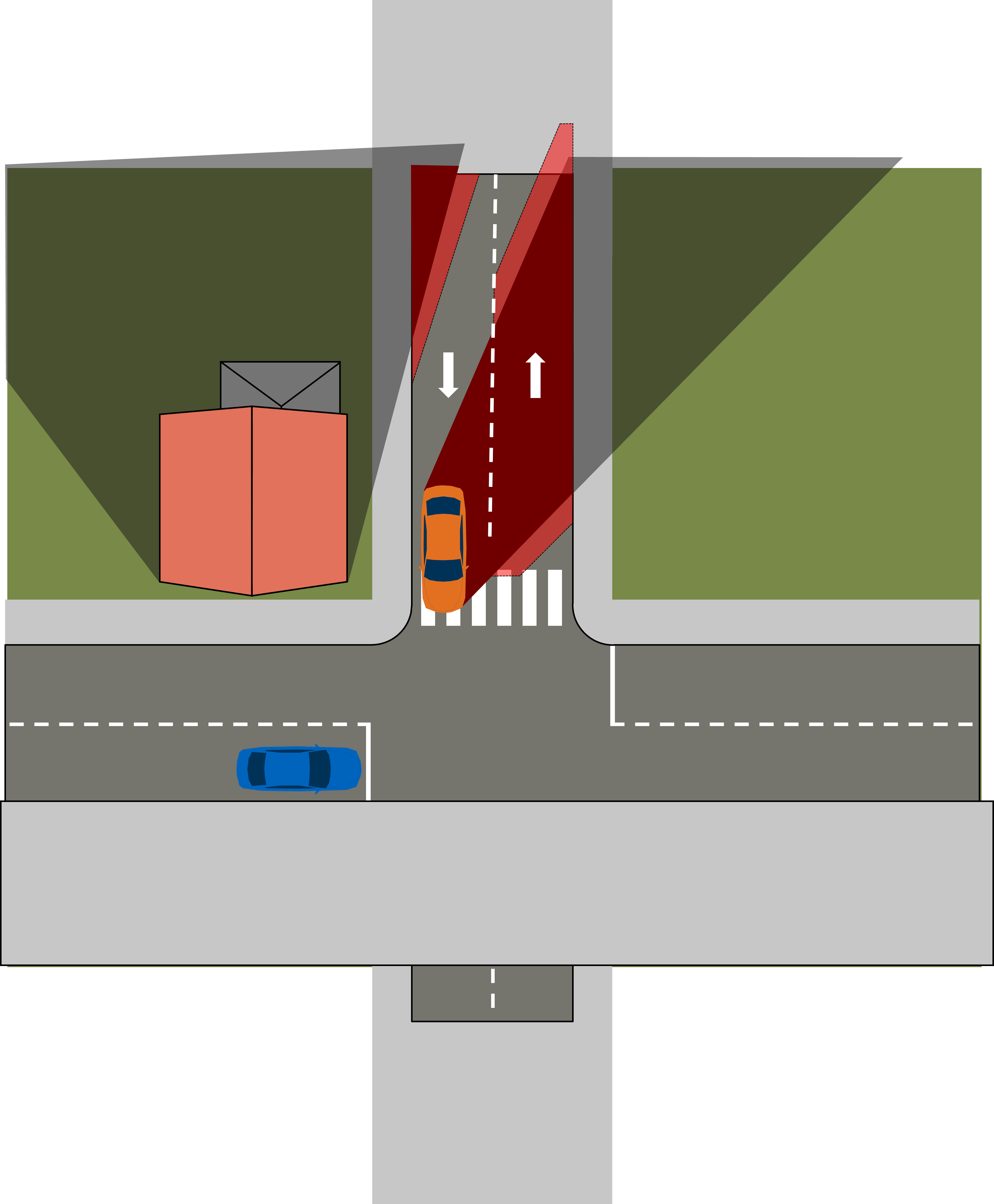}}} \\
    \subfloat[][Sensor measurements update the propagated occlusions, refining the occluded areas $\mathcal{A}_\mathrm{o}^\ell$]{\label{fig:tracking_1} \fbox{\includegraphics[width=0.38\textwidth, trim={2cm 13cm 3cm 8.5cm},clip]{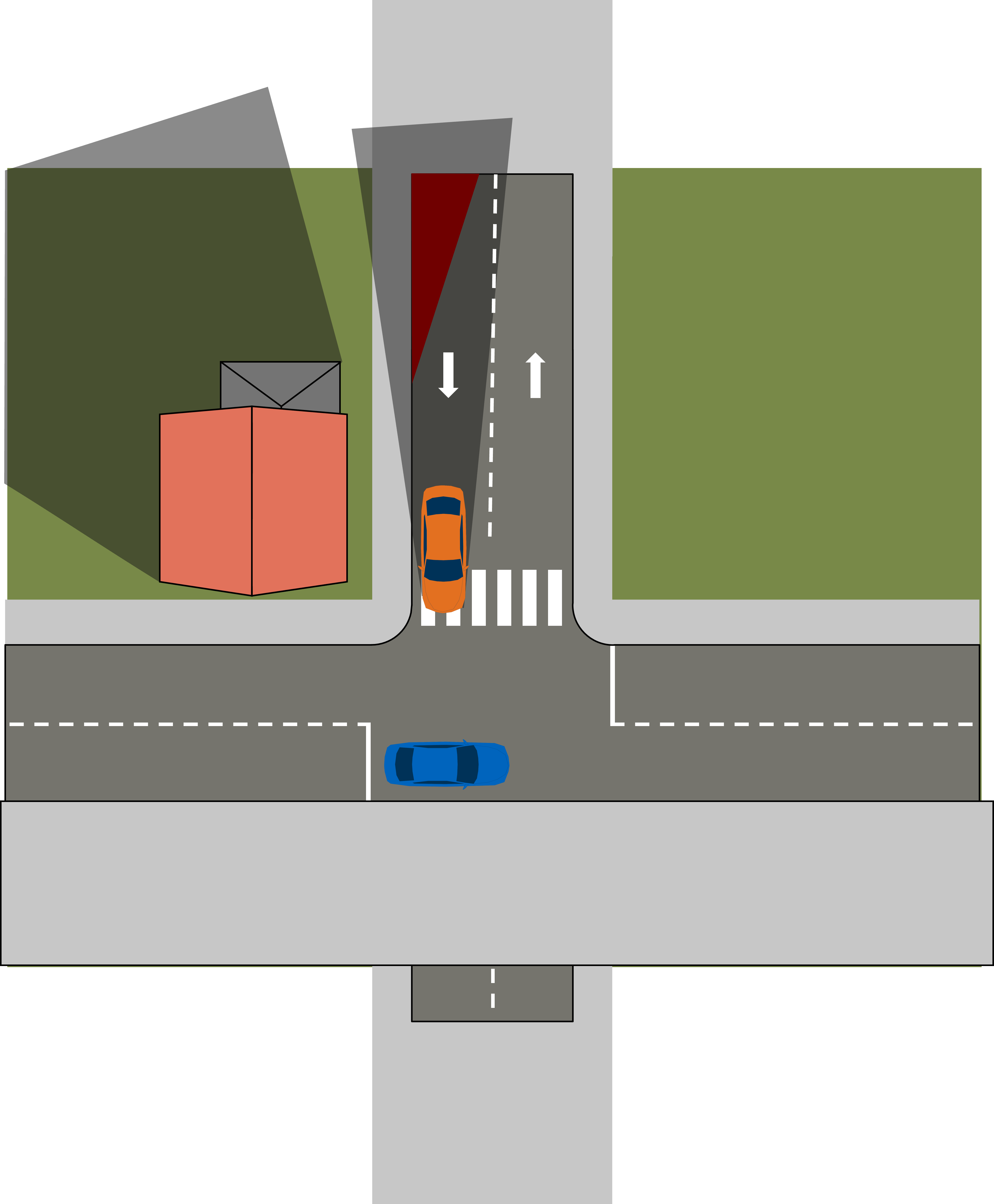}}}
    \caption{Illustration of an exemplary occlusion tracking process for multiple lanes $\ell \in \mathcal{L}$ at two timesteps. The visible area $\mathcal{A}_\mathrm{v}^\mathcal{L}$ is shown alongside the occluded areas $\mathcal{A}_\mathrm{o}^\ell$ (dark red) and their future propagation (light red). Grey shading represents currently unseen areas.}
    \label{fig:OcclusionTracking}%
\end{figure}

For subsequent updates, the occluded areas $\mathcal{A}_\mathrm{o}^\ell$ are individually propagated using a point-mass motion model, which projects the previously occluded states into the future based on the motion constraints $\mathcal{C}^\ell$. This projection captures the potential movement of hidden agents while ensuring that their dynamics remain valid within the lane's state space $\mathcal{X}_v^\ell$. After propagation, only those projected states that remain outside the visible area $\mathcal{A}_\mathrm{v}^\ell$ are retained as part of the updated occluded area. The updated occluded area for a lane $\ell$ at a given timestep is thus computed as:

\begin{equation}
    \mathcal{A}_\mathrm{o}^\ell = \left\{ x' \;\middle|\; 
    \begin{aligned}
        &x' = f(x, \mathcal{C}^\ell), \; x \in \mathcal{A}_\mathrm{o}^\ell, \\
        &x' \in \mathcal{X}_v^\ell, \; x' \notin \mathcal{A}_\mathrm{v}^\ell
    \end{aligned}
    \right\},
\end{equation}
where $x' = f(x, \mathcal{C}^\ell)$ denotes the projected state of a point mass $x \in \mathcal{A}_\mathrm{o}^\ell$ after propagation according to $\mathcal{C}^\ell$. For vehicle lanes, the occluded areas are expanded only in the direction of permitted traffic flow, while on sidewalks, the occluded areas are allowed to expand bidirectionally along the path to reflect pedestrian movement. After propagating the occlusions for each lane, the overall occluded area $\mathcal{A}_\mathrm{o}^\mathcal{L}$ is updated according to \Cref{eq:A_o_L}. This process is illustrated in \Cref{fig:OcclusionTracking}, which visualizes the propagation of exemplary occluded areas.

\subsection{Object Prediction and Tracking}

The tracking of previously visible obstacles $\mathcal{O}_\mathrm{v}$ is another aspect of the proposed occlusion-awareness module, ensuring that real obstacles are consistently represented, even when they temporarily leave $\mathcal{A}_\mathrm{v}^{\mathcal{L}}$, as proposed by~\cite{Pang2023}. For faster road users like vehicles and cyclists, future states are predicted using Wale-Net~\cite{walenet}, a neural network-based model trained to infer upcoming states based on observed movement patterns. In contrast, pedestrian states are estimated using a constant velocity approach, which assumes linear motion over the prediction horizon based on their current state.

Predictions of future positions involve uncertainty, as the motion of obstacles cannot be precisely determined. Typically, predictions are corrected at each timestep through measurement updates, which provide new information about the obstacle's state. In the case of tracking previously visible obstacles $\mathcal{O}_\mathrm{v}$ that have exited $\mathcal{A}_\mathrm{v}^{\mathcal{L}}$, the uncertainty grows more significantly. Without further sensor updates, even the currently assumed state of the obstacle becomes increasingly uncertain over time.

The uncertainty is mathematically modeled using a bivariate normal distribution, which accounts for the positional variance in both the $x$- and $y$-coordinates. Formally, the predicted state $\mathbf{X}_\mathrm{pred}$ of an occluded obstacle $o$ at time $t$ is represented as:
\begin{equation}
    \mathbf{X}_\mathrm{pred}(t) \sim \mathcal{N}\big(\mathbf{\mu}(t), \mathbf{\Sigma}(t)\big),
\end{equation}
where $\mathbf{\mu}(t)$ is the mean position of the predicted state, and $\mathbf{\Sigma}(t)$ is the covariance matrix describing the uncertainty, which increases over time $t$. 

Tracking obstacles, even under growing uncertainty, ensures that the occlusion-aware motion planner anticipates their potential reappearance, reducing the risk of being unprepared for their return. If an obstacle remains unobserved for a predefined duration, it is removed from the tracked obstacles.


\subsection{Phantom Agent Generation and Prediction Evaluation}

The generation of PAs builds on the spawn point creation process proposed in \cite{Moller2024}, extending it to allow for more flexible spawn point placement and the inclusion of diverse speed and acceleration profiles. PAs are generated to represent potential road users in occluded areas $\mathcal{A}_\mathrm{o}^\mathcal{L}$ identified during occlusion tracking.

Spawn points $\mathcal{X}_\mathrm{SP}$ are identified within $\mathcal{A}_\mathrm{o}^\mathcal{L}$. A spawn point \(x_\mathrm{SP} \in \mathcal{X}_\mathrm{SP}\) is considered valid if it lies within occluded area $\mathcal{A}_\mathrm{o}^{\mathcal{L}}$ and does not intersect visible obstacles $\mathcal{O}_\mathrm{v}$:
\begin{equation}
    \mathcal{X}_\mathrm{SP} = \{ x_\mathrm{SP} \;|\; x_\mathrm{SP} \in \mathcal{A}_\mathrm{o}^{\mathcal{L}} \; \land \; x_\mathrm{SP} \notin \mathcal{O}_\mathrm{v} \}.
\end{equation}
For pedestrians, static spawn points are placed behind static obstacles (e.g., buildings, parked vehicles) where occlusions are most critical. The algorithm assumes a worst-case prediction where pedestrians move directly toward $\Gamma$.

Dynamic spawn points $\mathcal{X}_\mathrm{SP}^\mathrm{dyn}$ for cyclists and vehicles are identified in occluded areas $\mathcal{A}_\mathrm{o}^\mathcal{L}$ near $\Gamma$. These points are further filtered to include only locations where potential paths $\gamma$ may intersect with the EV's driving corridor:
\begin{equation}
    \mathcal{X}_\mathrm{SP}^\mathrm{dyn} = \{ x_\mathrm{SP} \;|\; x_\mathrm{SP} \in \mathcal{A}_\mathrm{o}^\mathcal{L} \; \land \; \gamma(x_\mathrm{SP}) \cap \Gamma \neq \emptyset \}.
\end{equation}

For each spawn point \(x_\mathrm{SP} \in \mathcal{X}_\mathrm{SP}\), multiple predictions are generated to account for the diverse dynamics of different road users. Each prediction is expressed as:
\begin{equation}
    \mathbf{X}_\mathrm{pred}(t) = f(x_\mathrm{SP}, v, a, t), \quad t \in [t_0, t_0 + T_\mathrm{pred}],
\end{equation}
where \(x_\mathrm{SP}\) is the spawn point, \(v\) the initial speed, \(a\) the acceleration (or deceleration), and \(T_\mathrm{pred}\) the prediction horizon.

These predictions are subsequently used in the trajectory safety assessment step. This step is based on the approach proposed in \cite{Moller2024}, with a particular focus on evaluating collision risks and potential damage. The explicit consideration of differing PA characteristics (e.g., VRU vs. non-VRU) allows for a nuanced risk assessment that goes beyond simple metrics such as time-to-collision (TTC). For instance, while TTC focuses solely on the temporal aspect of a potential collision, it does not account for the varying severity of a collision depending on the type of road user involved.

While our method does not guarantee complete collision avoidance, it ensures a balance between safety and efficiency. Overly conservative behavior, such as assuming every occluded area is fully occupied by worst-case scenarios, is avoided. Instead, the occlusion-awareness module evaluates the generated predictions to prioritize responses based on the likelihood and severity of potential collisions.

\section{Results \& Analysis}
\label{sec:results}

Our proposed methodology is evaluated in the 2D simulation environment CommonRoad~\cite{commonroad}, using an open-source motion planner~\cite{frenetix}. The planner generates multiple trajectory samples, selecting the optimal trajectory based on weighted cost functions. Our occlusion-awareness module is integrated to extend the planner by incorporating an additional safety assessment step, as shown in \Cref{fig:framework}.

Evaluation scenarios focus on conditions where real-world accidents frequently occur, such as intersections with obstructed visibility~\cite{NHTSADemographicTrends, Olszewski2019}. The velocity profiles, risk progression, and total occluded areas $\mathcal{A}_\mathrm{o}^\mathcal{L}$ are analyzed to demonstrate the benefits of our approach.

\subsection{Velocity and Acceleration Profiles with Obstacle Tracking}

\begin{figure}[!ht]
    \centering
    %
    \newcommand{\plotheight}{3.2cm}
\newcommand{\xmin}{0}
\newcommand{\xmax}{70}

\begin{tikzpicture}[font=\footnotesize]


\begin{axis}[
/pgf/number format/.cd,
1000 sep={},
height=\plotheight,
width=6.7cm,
scale only axis,
scaled ticks=false,
scaled ticks=false,
tick label style={/pgf/number format/fixed},
xlabel={Timestep $t$ in \si{\second}},
ylabel={Velocity $v$ in \si{\meter\per\second}},
xmin=\xmin, xmax=\xmax,
ymin=0, ymax=12.5,
axis y line*=left,
ytick = {0, 3, ..., 12}
]

\addplot[forget plot, draw=none, fill=Grey, fill opacity=0.7, line width=5pt] coordinates {(3,-1) (3, 13) (18,13) (18,-1)};
\label{plot:vehicle-is-visible-again}

\addplot [ultra thick, black]
table {%
0 11
1 11.0115
2 11.0436
3 11.0792
4 11.1014
5 11.1096
6 11.1105
7 11.0894
8 11.0319
9 10.9422
10 10.8282
11 10.6037
12 10.2886
13 10.0735
14 9.79144
15 9.45584
16 9.24688
17 8.98518
18 8.68167
19 8.49705
20 8.26725
21 7.99721
22 7.8306
23 7.62245
24 7.37733
25 7.22044
26 7.01671
27 6.77333
28 6.60785
29 6.38004
30 6.10187
31 5.90506
32 5.62792
33 5.29132
34 5.17081
35 5.16114
36 5.25101
37 5.36849
38 5.5767
39 5.86303
40 6.04961
41 6.2841
42 6.56029
43 6.73291
44 6.951
45 7.20733
46 7.36041
47 7.54394
48 7.75384
49 7.87763
50 8.02557
51 8.19495
52 8.29704
53 8.42134
54 8.56417
55 8.65268
56 8.76475
57 8.89667
58 8.9798
59 9.08612
60 9.21198
61 9.28722
62 9.37716
63 9.47979
64 9.54002
65 9.61155
66 9.69308
67 9.73944
68 9.79212
69 9.85057
};
\label{plot:velocity-tracking}

\addplot [ultra thick, black, dashed]
table {%
0 11
1 11.0115
2 11.0436
3 11.0792
4 11.1014
5 11.1096
6 11.1105
7 11.1085
8 11.1057
9 11.1027
10 11.0816
11 11.0243
12 10.935
13 10.8691
14 10.7778
15 10.6642
16 10.5757
17 10.4378
18 10.2563
19 10.1076
20 9.86884
21 9.54913
22 9.28372
23 8.86393
24 8.31784
25 7.96393
26 7.51453
27 6.98915
28 6.67072
29 6.2819
30 5.83725
31 5.54186
32 5.14157
33 4.66455
34 4.46449
35 4.36504
36 4.3577
37 4.43062
38 4.60493
39 4.86702
40 5.05759
41 5.32107
42 5.64656
43 5.85984
44 6.14035
45 6.47762
46 6.67954
47 6.91974
48 7.19181
49 7.35108
50 7.54028
51 7.75524
52 7.8822
53 8.03256
54 8.20343
55 8.30598
56 8.43035
57 8.57312
58 8.66165
59 8.77383
60 8.90586
61 8.98914
62 9.09553
63 9.22139
64 9.29661
65 9.38638
66 9.48871
67 9.54868
68 9.62001
69 9.70146
70 9.74787
};
\label{plot:velocity-no-tracking}

\end{axis}


\begin{axis}[
/pgf/number format/.cd,
1000 sep={},
height=\plotheight,
width=6.7cm,
legend style={
	at={(0.5,-0.25)}, 
	anchor=north,
	legend columns=2,
	cells={anchor=center},
	draw=none,
	column sep=0.25em,
	row sep=0.1em,
},
scale only axis,
scaled ticks=false,
scaled ticks=false,
tick label style={/pgf/number format/fixed},
ylabel={Acceleration $a$ in \si{\meter\per\square\second}},
ylabel style={color=TUMBlue},
xmin=\xmin, xmax=\xmax,
ymin=-3, 
ymax=3,
axis x line*=none,
axis y line*=right,
xtick = {-10},
ylabel shift={-6pt}
]

\addplot [ultra thick, TUMBlue]
table {%
0 -0.049346
1 0.18101
2 0.306199
3 0.247909
4 0.127634
5 0.0262802
6 -0.0122972
7 -0.204518
8 -0.371791
9 -0.259852
10 -0.862416
11 -1.36542
12 -0.884794
13 -1.25429
14 -1.55547
15 -0.894575
16 -1.18562
17 -1.42192
18 -0.801857
19 -1.04038
20 -1.25422
21 -0.722066
22 -0.941162
23 -1.13713
24 -0.655002
25 -0.907208
26 -1.1234
27 -0.651324
28 -0.992873
29 -1.27466
30 -0.748294
31 -1.20229
32 -1.55197
33 -0.898804
34 -0.316364
35 0.209555
36 0.339076
37 0.824333
38 1.2465
39 0.802483
40 1.05762
41 1.28169
42 0.737208
43 0.982625
44 1.19177
45 0.681961
46 0.84445
47 0.986563
48 0.55401
49 0.682054
50 0.796671
51 0.449924
52 0.570391
53 0.671919
54 0.378081
55 0.504816
56 0.613215
57 0.351764
58 0.476875
59 0.583707
60 0.33632
61 0.414815
62 0.483237
63 0.270818
64 0.330428
65 0.383528
66 0.214892
67 0.247874
68 0.278218
69 0.15291
};
\label{plot:acceleration-tracking}

\addplot [ultra thick, TUMBlue, dashed]
table {%
0 -0.049346
1 0.18101
2 0.306199
3 0.247909
4 0.127634
5 0.0262802
6 -0.0122972
7 -0.0169715
8 -0.0152372
9 -0.00637071
10 -0.198956
11 -0.370278
12 -0.259905
13 -0.395854
14 -0.515098
15 -0.308741
16 -0.570401
17 -0.802883
18 -0.503796
19 -0.977717
20 -1.40616
21 -0.895975
22 -1.73702
23 -2.4364
24 -1.4961
25 -2.02144
26 -2.44945
27 -1.38673
28 -1.77669
29 -2.09372
30 -1.16692
31 -1.76311
32 -2.2183
33 -1.26803
34 -0.744103
35 -0.262444
36 0.0858259
37 0.626804
38 1.09982
39 0.750179
40 1.14208
41 1.47936
42 0.879231
43 1.24065
44 1.55003
45 0.899967
46 1.10707
47 1.28325
48 0.714227
49 0.877826
50 1.02053
51 0.579198
52 0.703897
53 0.811455
54 0.456019
55 0.571864
56 0.671475
57 0.377993
58 0.50603
59 0.615075
60 0.353804
61 0.478534
62 0.58481
63 0.336941
64 0.414304
65 0.481396
66 0.268525
67 0.328416
68 0.382573
69 0.215234
70 0.248532
};
\label{plot:acceleration-no-tracking}

\legend{}

\end{axis}


\begin{axis}[
    height=\plotheight,  
    width=7cm,
    hide axis,
    axis x line=none,
    axis y line=none,
    tick style=none,
    enlarge x limits=false,
    enlarge y limits=false,
    xmin=0, xmax=1,
    ymin=0, ymax=1,
    legend columns=2,
    legend style=
    {
        at={(0.65,-0.5)}, 
        anchor=north,
        legend columns=2,
        cells={anchor=center},
        draw=none,
        column sep=0.25em,
        row sep=0.1em,
    },
]
\addlegendimage{ultra thick, black}
\addlegendentry{w/ tracking ($v$)}

\addlegendimage{ultra thick, TUMBlue}
\addlegendentry{w/ tracking ($a$)}

\addlegendimage{ultra thick, black, dashed}
\addlegendentry{w/o tracking ($v$)}

\addlegendimage{ultra thick, TUMBlue, dashed}
\addlegendentry{w/o tracking ($a$)}


\end{axis}

\end{tikzpicture}%

    \vspace{-0.3cm}
    \caption{Velocity and acceleration profiles for a scenario where another vehicle is occluded for a period of time (shaded area). The plot illustrates the differences in profiles when the occluded vehicle is tracked versus when it is not tracked during the occlusion period.}
    \label{fig:results_vehicle_tracking}
\end{figure}
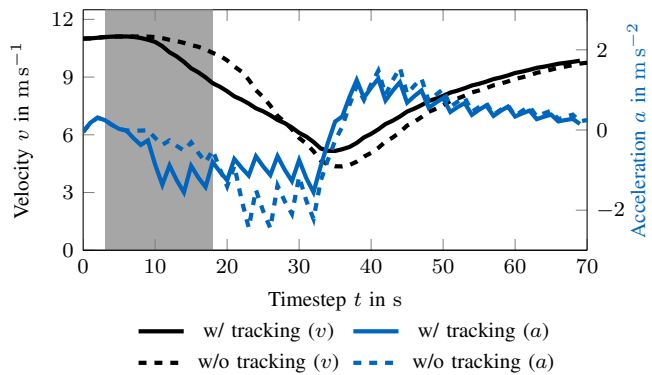

\Cref{fig:results_vehicle_tracking} presents the velocity and acceleration profiles for an intersection scenario where a vehicle becomes occluded for a period of time (shaded area). An exemplary situation marks a scenario, where a vehicle is temporarily hidden behind a static obstacle, such as a building. The comparison highlights two cases: one where the occluded vehicle is continuously tracked and one where it is not.
In the tracking case, the occlusion-aware motion planner anticipates the reappearance of the occluded vehicle and begins decelerating earlier—before the vehicle becomes visible again. This proactive adjustment leads to smoother velocity and acceleration profiles. In contrast, when tracking is disabled, the planner reacts only after the vehicle re-enters the visible area, resulting in abrupt deceleration and a lower minimum velocity as both vehicles are already closer to the intersection. Similar to findings in \cite{Pang2023}, tracking leads to more consistent trajectories with fewer abrupt decisions. Additionally, safety is improved, as tracking ensures greater separation from the other vehicle and allows the EV to approach the intersection at a lower, more controllable speed, leaving it better prepared to stop if necessary.

\subsection{Qualitative Analysis of a T-Junction Scenario}

\begin{figure}[!ht]
    \centering
    \hspace{1mm}
    \begin{tikzpicture}[font=\scriptsize]
        \node[inner sep=0pt] at (0.6,0) {\includegraphics[height=2.5mm]{input/ego.png}};
        \node[align=left, anchor=west] at (0.9,0) {ego \\ vehicle};
    
        \node[inner sep=0pt] at (2.3,0) {\includegraphics[height=2.5mm]{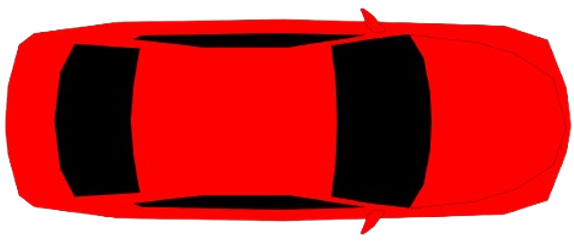}};
        \node[align=left, anchor=west] at (2.6,0) {static \\ obstacle};
    
        \draw[thick, black, line width=1pt] (3.8,0) -- (4.1,0);
        \node[align=left, anchor=west] at (4.2, 0) {trajectory};

        \node[inner sep=0pt] at (5.6,0) {\phantomPrediction};
        \node[align=left, anchor=west] at (5.8,0) {PA \\ prediction};

        \node[inner sep=0pt] at (7.3,0) {\phantomBike};
        \node[inner sep=0pt] at (7.8,0) {\phantomPedestrian};
        \node[align=left, anchor=west] at (8.0,0) {PAs};
    \end{tikzpicture}
    \subfloat[][Timestep 0]{\label{fig:qualitative_0} \fbox{\includegraphics[width=0.4\textwidth, trim={9.5cm 4.5cm 9.5cm 8.5cm},clip]{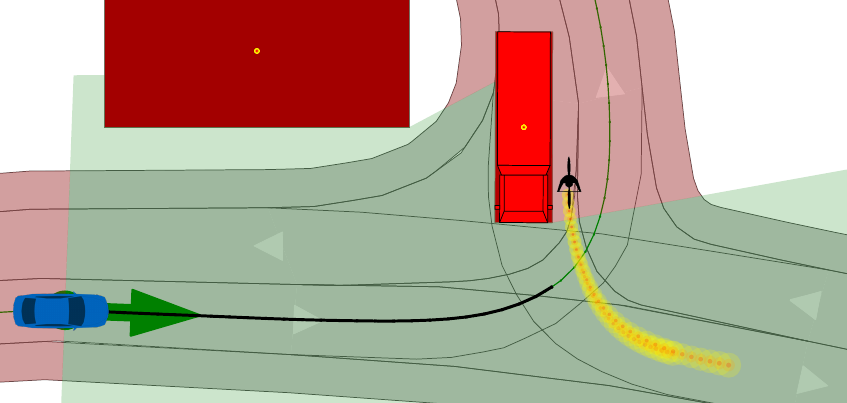}}} \\
    \subfloat[][Timestep 40]{\label{fig:qualitative_40} \fbox{\includegraphics[width=0.4\textwidth, trim={9.5cm 7.5cm 9.5cm 5.5cm},clip]{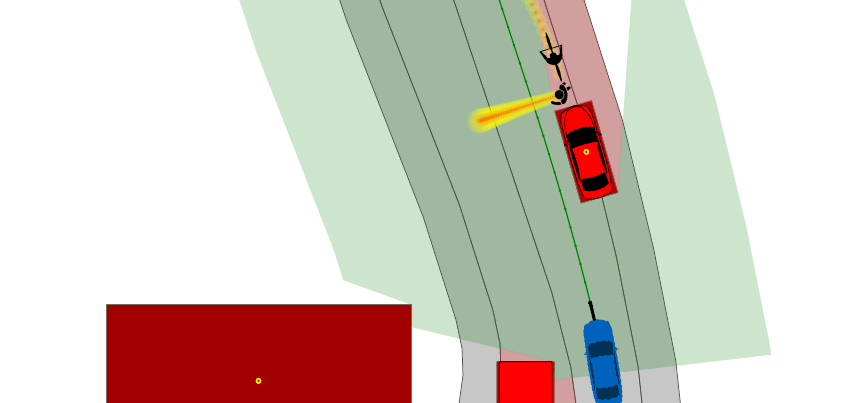}}}
    \caption{Visualization of a T-junction scenario with occlusion-aware planning. The stationary truck and the parked car create persistent occlusions. PAs are generated in critical areas.}
    \label{fig:QualitativeEvaluation}%
\end{figure}

In the T-junction scenario depicted in \Cref{fig:QualitativeEvaluation}, a stationary truck blocks the view into the intersection, creating persistent occlusions. 
PAs are placed in critical areas, such as behind the truck and along paths that intersect with the EV’s planned trajectory. Their predictions, including selected speed profiles, are visualized to highlight the safety-relevant agents. For clarity, not all generated profiles are displayed. This scenario serves as a foundation for further analyses in the following sections.

\subsection{Sensitivity Analysis of Risk Thresholds}

\begin{figure}[!ht]
    \centering
    \hspace{-0.3cm}
    \begin{tikzpicture}[font=\footnotesize]

\begin{axis}[
name=s_v_plot,
/pgf/number format/.cd,
1000 sep={},
height=3.0cm,
width=7.5cm,
legend style={
	at={(0.5,-0.25)}, 
	anchor=north,
	legend columns=2,
	cells={anchor=center},
	draw=none,
	column sep=0.25em,
	row sep=0.1em,
},
scale only axis,
scaled ticks=false,
scaled ticks=false,
ylabel={Velocity $v$ in \si{\meter\per\second}},
x label style={at={(0.5,-0.1)},anchor=north},
xmin=157, xmax=215,
ymin=0, ymax=9.9,
ymajorgrids=true
]
\addplot [ultra thick, black]
table {%
156.119 7.00003
156.819 7.05934
157.531 7.22233
158.265 7.46695
159.019 7.62189
159.79 7.80638
160.581 8.01357
161.388 8.13252
162.208 8.27043
163.043 8.42387
163.889 8.49903
164.742 8.56506
165.602 8.62263
166.465 8.65003
167.332 8.67834
168.201 8.70715
169.072 8.69852
169.939 8.64599
170.799 8.55321
171.65 8.45408
172.487 8.2758
173.303 8.02938
174.097 7.81812
174.862 7.47668
175.588 7.0284
176.277 6.71298
176.927 6.27615
177.529 5.7411
178.085 5.36917
178.598 4.85487
179.053 4.2357
179.459 3.86469
179.824 3.42539
180.142 2.93962
180.423 2.66385
180.674 2.34764
180.891 2.00505
181.083 1.81329
181.253 1.59535
181.401 1.36061
181.534 1.31164
181.668 1.39374
181.816 1.59038
181.982 1.74416
182.167 1.95257
182.375 2.20886
182.603 2.35434
182.846 2.51145
183.105 2.67811
183.378 2.76905
183.66 2.86983
183.952 2.97845
184.252 3.03169
184.558 3.07969
184.868 3.12268
185.182 3.14318
185.497 3.16388
185.814 3.18452
186.133 3.19337
186.453 3.19961
186.773 3.20355
187.093 3.20177
187.413 3.1938
187.732 3.18008
188.05 3.16734
188.365 3.14614
188.678 3.11731
188.989 3.09565
189.297 3.06354
189.602 3.02177
189.902 2.99346
190.2 2.9554
190.493 2.90837
190.784 2.93223
191.083 3.04709
191.397 3.24102
191.727 3.37149
192.072 3.5312
192.434 3.71691
192.811 3.82594
193.2 3.95395
193.602 4.09872
194.016 4.18241
194.439 4.27978
194.873 4.38887
195.314 4.44629
195.762 4.50393
196.215 4.5613
196.673 4.58797
197.133 4.61103
197.595 4.63065
198.058 4.63578
198.522 4.63356
198.985 4.62458
199.447 4.61227
199.907 4.58811
200.364 4.55295
200.817 4.51535
201.266 4.44588
201.706 4.34895
202.137 4.282
202.561 4.19197
202.975 4.08129
203.38 3.99961
203.774 3.87859
204.154 3.72395
204.525 3.7062
204.9 3.81197
205.291 4.02546
205.701 4.19888
206.133 4.44953
206.593 4.76607
207.079 4.96376
207.587 5.2071
208.122 5.49005
208.679 5.66659
209.257 5.89047
209.859 6.15419
210.482 6.31196
211.122 6.50126
211.783 6.71781
212.461 6.84167
213.152 6.9833
213.858 7.14085
214.577 7.23909
215.307 7.36594
216.051 7.51694
216.807 7.61279
217.574 7.73587
218.355 7.88194
219.147 7.96938
219.949 8.07399
220.763 8.19344
221.585 8.26359
222.416 8.34694
223.255 8.44198
224.102 8.49605
224.955 8.55747
225.814 8.62559
226.678 8.66591
227.547 8.71513
228.422 8.77254
229.3 8.80807
230.183 8.85291
231.071 8.90632
};
\addlegendentry{$R_\mathrm{max} = 0.05$}
\addplot [ultra thick, TUMBlue]
table {%
156.119 7.00003
156.819 7.05934
157.531 7.22233
158.265 7.46695
159.019 7.62189
159.79 7.80638
160.581 8.01357
161.388 8.13252
162.208 8.27043
163.043 8.42387
163.889 8.49903
164.742 8.56506
165.602 8.62263
166.465 8.65003
167.332 8.67834
168.201 8.70715
169.072 8.70231
169.94 8.66101
170.803 8.58672
171.657 8.47489
172.494 8.24695
173.303 7.92962
174.085 7.67869
174.835 7.30027
175.542 6.82622
176.21 6.52921
176.845 6.15955
177.44 5.7324
177.999 5.41257
178.517 4.931
178.981 4.32823
179.396 3.95889
179.77 3.51605
180.098 3.0226
180.386 2.74101
180.644 2.41709
180.869 2.06542
181.066 1.86829
181.242 1.64404
181.394 1.40234
181.53 1.3502
181.668 1.4299
181.82 1.62482
181.99 1.79157
182.181 2.03714
182.4 2.35005
182.644 2.52461
182.905 2.70483
183.185 2.88971
183.479 2.98901
183.783 3.09911
184.099 3.21861
184.424 3.28604
184.757 3.36338
185.097 3.44885
185.444 3.49041
185.795 3.52636
186.149 3.55698
186.506 3.5701
186.863 3.58145
187.222 3.59111
187.581 3.59117
187.94 3.58285
188.298 3.56675
188.654 3.55347
189.008 3.53375
189.36 3.50807
189.71 3.49087
190.058 3.46806
190.404 3.44006
190.748 3.47247
191.101 3.59243
191.469 3.7884
191.855 3.91877
192.254 4.07763
192.671 4.2618
193.102 4.37368
193.547 4.51129
194.005 4.67123
194.477 4.76488
194.959 4.87406
195.452 4.99602
195.955 5.05679
196.464 5.11184
196.977 5.16142
197.494 5.17987
198.013 5.18869
198.532 5.18852
199.05 5.18176
199.568 5.16643
200.083 5.14324
200.597 5.12281
201.108 5.08902
201.614 5.04286
202.117 5.00616
202.615 4.94942
203.106 4.87464
203.591 4.82106
204.07 4.74522
204.54 4.64984
205.004 4.66101
205.476 4.7887
205.965 5.01778
206.475 5.18343
207.004 5.40131
207.557 5.66483
208.131 5.8338
208.725 6.05149
209.343 6.31028
209.981 6.46595
210.637 6.65329
211.313 6.86802
212.006 6.99444
212.712 7.14406
213.435 7.31291
214.171 7.41614
214.919 7.54588
215.681 7.69791
216.456 7.79337
217.241 7.91523
218.039 8.05934
218.849 8.14237
219.668 8.2367
220.497 8.34092
221.334 8.39897
222.177 8.464
223.027 8.53537
223.883 8.5772
224.743 8.62789
225.608 8.6867
226.479 8.72294
227.353 8.76854
228.233 8.82275
229.117 8.85684
230.004 8.90036
230.897 8.95255
231.794 8.98561
232.694 9.02801
233.6 9.07902
234.509 9.10619
235.421 9.1324
236.335 9.15731
237.252 9.1696
238.169 9.18223
239.088 9.19507
};
\addlegendentry{$R_\mathrm{max} = 0.10$}
\addplot [ultra thick, TUMOrange]
table {%
156.119 7.00003
156.819 7.05934
157.531 7.22233
158.265 7.46695
159.019 7.62189
159.79 7.80638
160.581 8.01357
161.388 8.13252
162.208 8.27043
163.043 8.42387
163.889 8.49903
164.742 8.56506
165.602 8.62263
166.465 8.65003
167.332 8.67834
168.201 8.70715
169.072 8.71586
169.944 8.71346
170.814 8.70075
171.683 8.6602
172.544 8.56148
173.394 8.41242
174.229 8.28784
175.049 8.09241
175.846 7.83623
176.621 7.64939
177.373 7.38216
178.095 7.04963
178.789 6.81108
179.454 6.47292
180.081 6.06097
180.675 5.82025
181.244 5.5428
181.783 5.23525
182.3 5.12331
182.811 5.10076
183.323 5.15695
183.841 5.2019
184.363 5.24446
184.89 5.28422
185.419 5.30351
185.951 5.32298
186.484 5.34243
187.019 5.35307
187.555 5.36537
188.092 5.37902
188.63 5.38718
189.169 5.39723
189.71 5.40884
190.251 5.4299
190.796 5.47833
191.348 5.55042
191.906 5.62871
192.476 5.77321
193.063 5.97238
193.665 6.08451
194.279 6.19757
194.905 6.31056
195.539 6.36318
196.177 6.40867
196.82 6.4474
197.466 6.46397
198.113 6.4783
198.761 6.4905
199.411 6.49451
200.06 6.49552
200.71 6.49372
201.359 6.49104
202.008 6.48626
202.656 6.47951
203.304 6.47611
203.951 6.47354
204.599 6.47171
205.247 6.50165
205.901 6.58873
206.566 6.72661
207.243 6.82341
207.932 6.95051
208.634 7.10468
209.35 7.20379
210.076 7.33148
210.817 7.48331
211.57 7.57959
212.333 7.70319
213.111 7.84981
213.9 7.93757
214.699 8.04257
215.509 8.16244
216.329 8.23284
217.156 8.31648
217.993 8.41184
218.836 8.4661
219.686 8.52773
220.542 8.59608
221.404 8.63654
222.27 8.68594
223.141 8.74354
224.017 8.7792
224.897 8.82419
225.782 8.87778
226.672 8.91154
227.565 8.95467
228.463 9.00643
229.365 9.03924
230.271 9.08134
231.182 9.13199
232.096 9.1581
233.013 9.18179
233.932 9.20319
234.853 9.2136
235.775 9.22445
236.698 9.2356
237.622 9.24189
238.547 9.24937
239.472 9.25784
240.398 9.263
};
\addlegendentry{$R_\mathrm{max} = 0.20$}
\addplot [ultra thick, black, dashed]
table {%
156.119 7.00003
156.819 7.05934
157.531 7.22233
158.265 7.46695
159.019 7.62189
159.79 7.80638
160.581 8.01357
161.388 8.13252
162.208 8.27043
163.043 8.42387
163.889 8.50917
164.745 8.6044
165.611 8.70832
166.484 8.76579
167.364 8.82975
168.25 8.89955
169.142 8.94025
170.038 8.98926
170.94 9.04594
171.846 9.08078
172.756 9.12466
173.671 9.1769
174.59 9.20356
175.512 9.22724
176.436 9.24809
177.361 9.25808
178.287 9.2683
179.214 9.27856
180.142 9.28414
181.071 9.29065
182 9.29795
182.93 9.30224
183.86 9.3076
184.791 9.3139
185.723 9.31776
186.655 9.32287
187.588 9.32904
188.521 9.33295
189.455 9.33791
190.389 9.34376
191.323 9.34746
192.258 9.35215
193.194 9.35764
194.13 9.36108
195.066 9.36543
196.003 9.37057
196.94 9.37381
197.878 9.37794
198.816 9.38282
199.754 9.38588
200.693 9.38977
201.632 9.39435
202.572 9.39723
203.512 9.40088
204.452 9.4052
205.393 9.40791
206.334 9.41135
207.275 9.41542
208.217 9.41797
209.159 9.42122
210.101 9.42505
211.043 9.42746
211.986 9.43052
212.93 9.43413
213.873 9.4364
214.817 9.43929
215.761 9.4427
216.705 9.44484
217.65 9.44756
218.595 9.45078
219.54 9.45279
220.485 9.45536
221.431 9.45839
222.377 9.4603
223.323 9.46272
224.27 9.46558
225.216 9.46737
226.163 9.46965
227.11 9.47235
228.058 9.47404
229.005 9.47619
229.953 9.47874
230.901 9.48033
231.849 9.48236
232.797 9.48476
233.746 9.48626
234.694 9.48818
235.643 9.49044
236.593 9.49186
237.542 9.49366
238.491 9.49579
239.441 9.49713
240.391 9.49883
};
\addlegendentry{$R_\mathrm{max} = \infty$}

\legend{};

\end{axis}



\begin{axis}[
name=s_risk_plot,
at={(s_v_plot.below south)},
yshift=-0.1cm,
anchor=north,
/pgf/number format/.cd,
1000 sep={},
height=3.2cm,
width=7.5cm,
legend style={
	at={(0.5,-0.25)}, 
	anchor=north,
	legend columns=2,
	cells={anchor=center},
	draw=none,
	column sep=0.25em,
	row sep=0.1em,
},
scale only axis,
scaled ticks=false,
scaled ticks=false,
tick label style={/pgf/number format/fixed},
xlabel={Coordinate $s$ in \si{\meter}},
ylabel={Risk $R$},
x label style={at={(0.5,-0.1)},anchor=north},
xmin=157, xmax=215,
ymin=0, ymax=0.59,
ytick = {0.10, 0.20, 0.30, 0.40, 0.50},
ymajorgrids=true,
]
\addplot [ultra thick, black]
table {%
156.119 0
156.819 0
157.531 0
158.265 0
159.019 0
159.79 0
160.581 0.000164965
161.388 0.000164965
162.208 0.000164965
163.043 0.00145017
163.889 0.00145017
164.742 0.00145017
165.602 0.00227405
166.465 0.00227405
167.332 0.00227405
168.201 0.033494
169.072 0.033494
169.939 0.033494
170.799 0.0301798
171.65 0.0301798
172.487 0.0301798
173.303 0.014489
174.097 0.014489
174.862 0.014489
175.588 0.01129
176.277 0.01129
176.927 0.01129
177.529 0.0326928
178.085 0.0326928
178.598 0.0326928
179.053 0.101939
179.459 0.101939
179.824 0.101939
180.142 0.141997
180.423 0.141997
180.674 0.141997
180.891 0.151153
181.083 0.151153
181.253 0.151153
181.401 0.000403734
181.534 0.000403734
181.668 0.000403734
181.816 0.0422225
181.982 0.0422225
182.167 0.0422225
182.375 0.0443641
182.603 0.0443641
182.846 0.0443641
183.105 0.0303449
183.378 0.0303449
183.66 0.0303449
183.952 0.00499731
184.252 0.00499731
184.558 0.00499731
184.868 0.0326391
185.182 0.0326391
185.497 0.0326391
185.814 0.0499363
186.133 0.0499363
186.453 0.0499363
186.773 0.0354367
187.093 0.0354367
187.413 0.0354367
187.732 0.049757
188.05 0.049757
188.365 0.049757
188.678 0.0314098
188.989 0.0314098
189.297 0.0314098
189.602 0.0492692
189.902 0.0492692
190.2 0.0492692
190.493 0.0333974
190.784 0.0333974
191.083 0.0333974
191.397 0.0496255
191.727 0.0496255
192.072 0.0496255
192.434 0.0495849
192.811 0.0495849
193.2 0.0495849
193.602 0.0450248
194.016 0.0450248
194.439 0.0450248
194.873 0.0315333
195.314 0.0315333
195.762 0.0315333
196.215 0.0413679
196.673 0.0413679
197.133 0.0413679
197.595 0.0477492
198.058 0.0477492
198.522 0.0477492
198.985 0.044922
199.447 0.044922
199.907 0.044922
200.364 0.0436773
200.817 0.0436773
201.266 0.0436773
201.706 0.038348
202.137 0.038348
202.561 0.038348
202.975 0.0437938
203.38 0.0437938
203.774 0.0437938
204.154 0
204.525 nan
204.9 nan
205.291 nan
205.701 nan
206.133 nan
206.593 nan
207.079 nan
207.587 nan
208.122 nan
208.679 nan
209.257 nan
209.859 nan
210.482 nan
211.122 nan
211.783 nan
212.461 nan
213.152 nan
213.858 nan
214.577 nan
215.307 nan
216.051 nan
216.807 nan
217.574 nan
218.355 nan
219.147 nan
219.949 nan
220.763 nan
221.585 nan
222.416 nan
223.255 nan
224.102 nan
224.955 nan
225.814 nan
226.678 nan
227.547 nan
228.422 nan
229.3 nan
230.183 nan
231.071 0
};
\addlegendentry{$R_\mathrm{max} = 0.05$}
\addplot [ultra thick, TUMBlue]
table {%
156.119 0
156.819 0
157.531 0
158.265 0
159.019 0
159.79 0
160.581 0.000164965
161.388 0.000164965
162.208 0.000164965
163.043 0.00145017
163.889 0.00145017
164.742 0.00145017
165.602 0.00227405
166.465 0.00227405
167.332 0.00227405
168.201 0.0912333
169.072 0.0912333
169.94 0.0912333
170.803 0.0989391
171.657 0.0989391
172.494 0.0989391
173.303 0.0774674
174.085 0.0774674
174.835 0.0774674
175.542 0.0914044
176.21 0.0914044
176.845 0.0914044
177.44 0.0303033
177.999 0.0303033
178.517 0.0303033
178.981 0.100173
179.396 0.100173
179.77 0.100173
180.098 0.137673
180.386 0.137673
180.644 0.137673
180.869 0.150506
181.066 0.150506
181.242 0.150506
181.394 0.0003835
181.53 0.0003835
181.668 0.0003835
181.82 0.0576525
181.99 0.0576525
182.181 0.0576525
182.4 0.0969023
182.644 0.0969023
182.905 0.0969023
183.185 0.0964547
183.479 0.0964547
183.783 0.0964547
184.099 0.0836847
184.424 0.0836847
184.757 0.0836847
185.097 0.0576949
185.444 0.0576949
185.795 0.0576949
186.149 0.0873546
186.506 0.0873546
186.863 0.0873546
187.222 0.0636772
187.581 0.0636772
187.94 0.0636772
188.298 0.0799127
188.654 0.0799127
189.008 0.0799127
189.36 0.0861024
189.71 0.0861024
190.058 0.0861024
190.404 0.0616855
190.748 0.0616855
191.101 0.0616855
191.469 0.0895369
191.855 0.0895369
192.254 0.0895369
192.671 0.0978254
193.102 0.0978254
193.547 0.0978254
194.005 0.0945433
194.477 0.0945433
194.959 0.0945433
195.452 0.0737182
195.955 0.0737182
196.464 0.0737182
196.977 0.0672497
197.494 0.0672497
198.013 0.0672497
198.532 0.0986151
199.05 0.0986151
199.568 0.0986151
200.083 0.0961533
200.597 0.0961533
201.108 0.0961533
201.614 0.0955634
202.117 0.0955634
202.615 0.0955634
203.106 0.0969281
203.591 0.0969281
204.07 0.0969281
204.54 0
205.004 nan
205.476 nan
205.965 nan
206.475 nan
207.004 nan
207.557 nan
208.131 nan
208.725 nan
209.343 nan
209.981 nan
210.637 nan
211.313 nan
212.006 nan
212.712 nan
213.435 nan
214.171 nan
214.919 nan
215.681 nan
216.456 nan
217.241 nan
218.039 nan
218.849 nan
219.668 nan
220.497 nan
221.334 nan
222.177 nan
223.027 nan
223.883 nan
224.743 nan
225.608 nan
226.479 nan
227.353 nan
228.233 nan
229.117 nan
230.004 nan
230.897 nan
231.794 nan
232.694 nan
233.6 nan
234.509 nan
235.421 nan
236.335 nan
237.252 nan
238.169 nan
239.088 0
};
\addlegendentry{$R_\mathrm{max} = 0.10$}
\addplot [ultra thick, TUMOrange]
table {%
156.119 0
156.819 0
157.531 0
158.265 0
159.019 0
159.79 0
160.581 0.000164965
161.388 0.000164965
162.208 0.000164965
163.043 0.00145017
163.889 0.00145017
164.742 0.00145017
165.602 0.00227405
166.465 0.00227405
167.332 0.00227405
168.201 0.194301
169.072 0.194301
169.944 0.194301
170.814 0.153571
171.683 0.153571
172.544 0.153571
173.394 0.175739
174.229 0.175739
175.049 0.175739
175.846 0.194576
176.621 0.194576
177.373 0.194576
178.095 0.198087
178.789 0.198087
179.454 0.198087
180.081 0.18798
180.675 0.18798
181.244 0.18798
181.783 0.0126684
182.3 0.0126684
182.811 0.0126684
183.323 0.151252
183.841 0.151252
184.363 0.151252
184.89 0.16928
185.419 0.16928
185.951 0.16928
186.484 0.173542
187.019 0.173542
187.555 0.173542
188.092 0.173996
188.63 0.173996
189.169 0.173996
189.71 0.198443
190.251 0.198443
190.796 0.198443
191.348 0.180989
191.906 0.180989
192.476 0.180989
193.063 0.159482
193.665 0.159482
194.279 0.159482
194.905 0.153729
195.539 0.153729
196.177 0.153729
196.82 0.191965
197.466 0.191965
198.113 0.191965
198.761 0.192828
199.411 0.192828
200.06 0.192828
200.71 0.194246
201.359 0.194246
202.008 0.194246
202.656 0.198943
203.304 0.198943
203.951 0.198943
204.599 0
205.247 nan
205.901 nan
206.566 nan
207.243 nan
207.932 nan
208.634 nan
209.35 nan
210.076 nan
210.817 nan
211.57 nan
212.333 nan
213.111 nan
213.9 nan
214.699 nan
215.509 nan
216.329 nan
217.156 nan
217.993 nan
218.836 nan
219.686 nan
220.542 nan
221.404 nan
222.27 nan
223.141 nan
224.017 nan
224.897 nan
225.782 nan
226.672 nan
227.565 nan
228.463 nan
229.365 nan
230.271 nan
231.182 nan
232.096 nan
233.013 nan
233.932 nan
234.853 nan
235.775 nan
236.698 nan
237.622 nan
238.547 nan
239.472 nan
240.398 0
};
\addlegendentry{$R_\mathrm{max} = 0.20$}
\addplot [ultra thick, black, dashed]
table {%
156.119 0
156.819 0
157.531 0
158.265 0
159.019 0
159.79 0
160.581 0.000164965
161.388 0.000164965
162.208 0.000164965
163.043 0.0162801
163.889 0.0162801
164.745 0.0162801
165.611 0.00840141
166.484 0.00840141
167.364 0.00840141
168.25 0.392216
169.142 0.392216
170.038 0.392216
170.94 0.4572
171.846 0.4572
172.756 0.4572
173.671 0.506375
174.59 0.506375
175.512 0.506375
176.436 0.564324
177.361 0.564324
178.287 0.564324
179.214 0.393767
180.142 0.393767
181.071 0.393767
182 0.338417
182.93 0.338417
183.86 0.338417
184.791 0.357018
185.723 0.357018
186.655 0.357018
187.588 0.267243
188.521 0.267243
189.455 0.267243
190.389 0.315342
191.323 0.315342
192.258 0.315342
193.194 0.349538
194.13 0.349538
195.066 0.349538
196.003 0.355485
196.94 0.355485
197.878 0.355485
198.816 0.353697
199.754 0.353697
200.693 0.353697
201.632 0.337594
202.572 0.337594
203.512 0.337594
204.452 0
205.393 nan
206.334 nan
207.275 nan
208.217 nan
209.159 nan
210.101 nan
211.043 nan
211.986 nan
212.93 nan
213.873 nan
214.817 nan
215.761 nan
216.705 nan
217.65 nan
218.595 nan
219.54 nan
220.485 nan
221.431 nan
222.377 nan
223.323 nan
224.27 nan
225.216 nan
226.163 nan
227.11 nan
228.058 nan
229.005 nan
229.953 nan
230.901 nan
231.849 nan
232.797 nan
233.746 nan
234.694 nan
235.643 nan
236.593 nan
237.542 nan
238.491 nan
239.441 nan
240.391 0
};
\addlegendentry{$R_\mathrm{max} = \infty$}

\end{axis}

    \end{tikzpicture}
    \vspace{-0.4cm}
    \caption{Velocity and risk profiles for the T-juntion scenario across four simulation runs with different risk thresholds $R_\mathrm{max}$.}
    \label{fig:result_sensitivity_velocity_risk}
\end{figure}
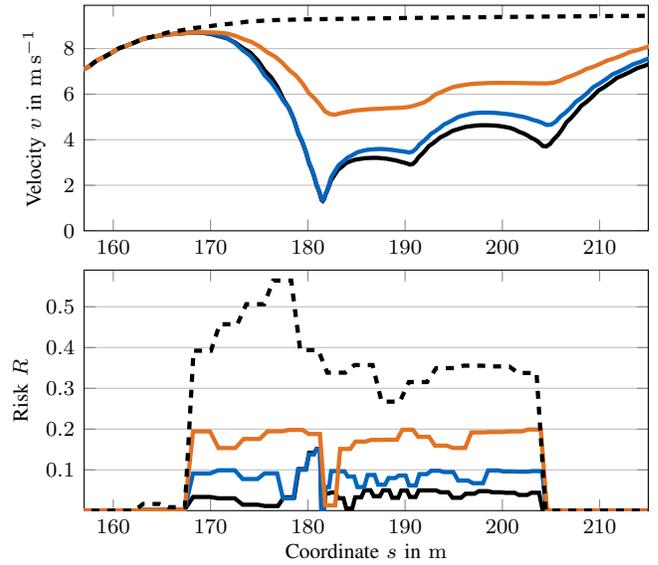

To evaluate the influence of the risk threshold \( R_\mathrm{max} \) on the planner's behavior, a sensitivity analysis is conducted. The threshold is varied between \( R_\mathrm{max} = 0.05 \) and \( R_\mathrm{max} = 0.20 \), alongside an unrestricted planner (\( R_\mathrm{max} = \infty \)). \Cref{fig:result_sensitivity_velocity_risk} visualizes velocity and respective risk profiles across the scenario progression.
Lower thresholds (stricter risk limits) lead to greater reductions in velocity to ensure compliance with the specified risk levels. This behavior is particularly evident during the turning maneuver at the T-junction, where the EV significantly slows down $(s=\SI{181}{\meter})$. In contrast, when passing the parked vehicle, less deceleration occurs due to the lower criticality of the situation $(s=\SI{204}{\meter})$.

An exception is observed at $s=\SI{180}{\meter}$, where the maximum risk for $R_\mathrm{max} \leq 0.10$ is briefly exceeded. This occurs because no other valid trajectory was available. The results show the planner's ability to adapt its behavior based on specified thresholds.

\subsection{Comparison of our Module with Other Approaches}

The performance of our proposed occlusion-awareness module is evaluated against three other approaches: a baseline planner without occlusion-awareness capabilities~\cite{frenetix}, the occlusion-aware planner from~\cite{Moller2024}, and an omniscient planner with complete environmental knowledge. The T-junction scenario (see \Cref{fig:QualitativeEvaluation}) is used for this comparison, with a real cyclist replacing the phantom cyclist at the position shown in \Cref{fig:qualitative_0}. This cyclist remains occluded by the truck until it becomes visible later in the scenario. 

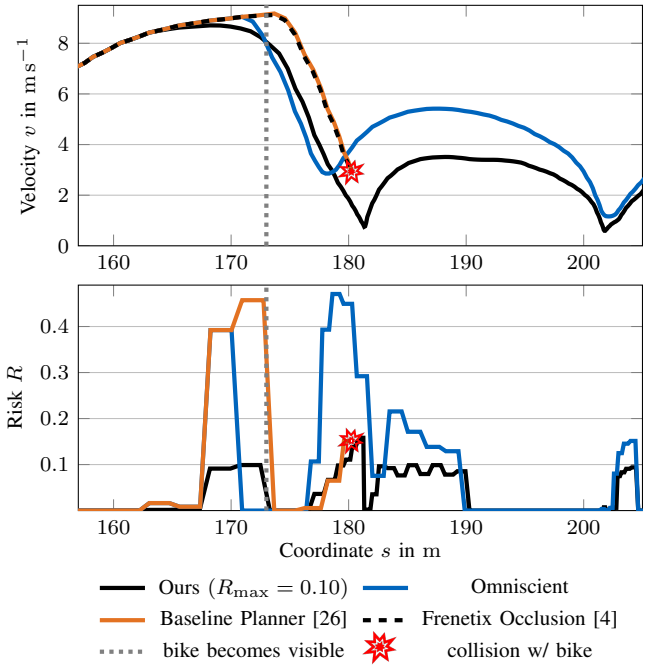
\begin{figure}[!ht]
    \centering
    \hspace{-0.3cm}
    \begin{tikzpicture}[font=\footnotesize]

\begin{axis}[
name=s_v_plot,
/pgf/number format/.cd,
1000 sep={},
height=3.2cm,
width=7.5cm,
legend style={
	at={(0.5,-0.25)}, 
	anchor=north,
	legend columns=2,
	cells={anchor=center},
	draw=none,
	column sep=0.25em,
	row sep=0.1em,
},
scale only axis,
scaled ticks=false,
scaled ticks=false,
ylabel={Velocity $v$ in \si{\meter\per\second}},
x label style={at={(0.5,-0.1)},anchor=north},
xmin=157, xmax=205,
ymin=0, ymax=9.5,
ymajorgrids=true
]
\addplot [ultra thick, black]
table {%
156.119 7.00003
156.819 7.05934
157.531 7.22233
158.265 7.46695
159.019 7.62189
159.79 7.80638
160.581 8.01357
161.388 8.13252
162.208 8.27043
163.043 8.42387
163.889 8.49903
164.742 8.56506
165.602 8.62263
166.465 8.65003
167.332 8.67834
168.201 8.70715
169.072 8.70231
169.94 8.66101
170.803 8.58672
171.657 8.47489
172.494 8.24695
173.303 7.92962
174.082 7.59904
174.814 7.00222
175.476 6.20188
176.072 5.69329
176.611 5.0715
177.083 4.37057
177.502 3.99694
177.882 3.61185
178.224 3.22523
178.536 3.01263
178.825 2.76443
179.088 2.48885
179.33 2.33713
179.555 2.17058
179.763 1.99228
179.958 1.88431
180.139 1.74745
180.306 1.58759
180.46 1.49143
180.604 1.37525
180.735 1.24339
180.856 1.16568
180.968 1.073
181.07 0.968677
181.164 0.907558
181.251 0.834958
181.33 0.753439
181.406 0.773083
181.489 0.90719
181.59 1.14083
181.712 1.29441
181.85 1.47491
182.008 1.67965
182.182 1.81413
182.372 1.99675
182.583 2.22025
182.811 2.35143
183.054 2.50188
183.312 2.66929
183.584 2.7671
183.866 2.88373
184.161 3.01658
184.466 3.09228
184.78 3.17799
185.102 3.27183
185.432 3.32132
185.767 3.37237
186.106 3.42474
186.45 3.44954
186.796 3.47109
187.144 3.48955
187.494 3.49748
187.844 3.50434
188.195 3.51018
188.546 3.50879
188.896 3.49962
189.245 3.48324
189.593 3.47016
189.939 3.45102
190.283 3.42629
190.625 3.41432
190.966 3.40557
191.306 3.39971
191.646 3.39803
191.986 3.39743
192.326 3.39776
192.665 3.3945
193.004 3.38429
193.342 3.36762
193.678 3.35485
194.013 3.33657
194.345 3.31318
194.675 3.28681
195.002 3.23626
195.322 3.16572
195.636 3.12206
195.946 3.07065
196.25 3.01215
196.55 2.97528
196.845 2.92719
197.135 2.86892
197.42 2.82316
197.699 2.75134
197.969 2.65721
198.232 2.59993
198.489 2.53272
198.738 2.4564
198.982 2.40601
199.219 2.33675
199.449 2.25057
199.671 2.19185
199.886 2.11146
200.092 2.01206
200.291 1.94416
200.48 1.85101
200.66 1.73639
200.83 1.6447
200.987 1.50209
201.129 1.32112
201.255 1.2094
201.37 1.07489
201.47 0.924637
201.558 0.838754
201.637 0.739859
201.706 0.632417
201.767 0.596085
201.826 0.59522
201.887 0.626686
201.951 0.654563
202.018 0.691152
202.09 0.73578
202.164 0.764294
202.243 0.801001
202.325 0.84528
202.41 0.870498
202.499 0.898227
202.59 0.928358
202.683 0.948214
202.779 0.976153
202.878 1.01161
202.98 1.04868
203.088 1.11934
203.204 1.22117
203.329 1.29159
203.461 1.38515
203.605 1.50001
203.757 1.57115
203.918 1.65889
204.088 1.76182
204.266 1.82264
204.451 1.8946
204.643 1.97673
204.843 2.04804
205.053 2.17168
205.278 2.33412
205.517 2.44265
205.768 2.57446
206.033 2.72785
206.311 2.82275
206.6 2.94299
206.902 3.08664
207.216 3.17753
207.54 3.29437
207.877 3.43514
208.226 3.5365
208.587 3.68518
208.966 3.87631
209.36 4.01387
209.772 4.2097
210.205 4.45688
210.659 4.63491
211.135 4.8894
211.639 5.20902
212.17 5.42376
212.726 5.70883
213.313 6.05027
213.927 6.2545
214.564 6.49743
215.227 6.77278
215.912 6.92697
216.613 7.09799
217.331 7.28254
218.065 7.39159
218.81 7.52585
219.571 7.68125
220.343 7.77801
221.127 7.90093
221.924 8.04589
222.733 8.13233
223.551 8.23555
224.381 8.35327
225.219 8.4255
};
\addlegendentry{Our Algorithm}
\addplot [ultra thick, TUMBlue]
table {%
156.119 7.00003
156.819 7.05934
157.531 7.22233
158.265 7.46695
159.019 7.62189
159.79 7.80638
160.581 8.01357
161.388 8.13252
162.208 8.27043
163.043 8.42387
163.889 8.50917
164.745 8.6044
165.611 8.70832
166.484 8.76579
167.364 8.82975
168.25 8.89955
169.142 8.94025
170.038 8.98926
170.94 9.04594
171.839 8.86017
172.7 8.29804
173.489 7.44257
174.205 6.86659
174.857 6.14168
175.43 5.31077
175.938 4.82773
176.393 4.26581
176.789 3.65133
177.139 3.35946
177.463 3.11414
177.764 2.91282
178.051 2.85436
178.337 2.85607
178.625 2.91322
178.92 3.00743
179.229 3.19803
179.562 3.47183
179.918 3.65383
180.295 3.88469
180.697 4.1581
181.12 4.31678
181.561 4.4984
182.021 4.70015
182.496 4.81121
182.983 4.93268
183.483 5.06336
183.993 5.1344
184.51 5.21207
185.036 5.29541
185.567 5.33635
186.103 5.37325
186.642 5.40633
187.183 5.41696
187.725 5.41801
188.267 5.41011
188.807 5.39576
189.345 5.366
189.88 5.32258
190.411 5.29106
190.938 5.24678
191.46 5.19074
191.977 5.14514
192.488 5.0721
192.99 4.97483
193.485 4.91169
193.972 4.8315
194.45 4.73566
194.921 4.66427
195.382 4.55619
195.831 4.4158
196.268 4.32813
196.696 4.22127
197.112 4.09684
197.518 4.0136
197.914 3.89872
198.297 3.7555
198.667 3.64882
199.025 3.49014
199.364 3.28699
199.686 3.14301
199.99 2.94123
200.272 2.69262
200.534 2.51903
200.774 2.27838
200.988 1.98828
201.178 1.81433
201.349 1.60827
201.499 1.38035
201.632 1.27979
201.756 1.21192
201.875 1.17437
201.992 1.16553
202.108 1.16209
202.224 1.16383
202.34 1.17058
202.458 1.18626
202.577 1.21028
202.698 1.24218
202.825 1.30839
202.959 1.40644
203.102 1.47927
203.254 1.58311
203.418 1.71548
203.592 1.80343
203.777 1.92012
203.975 2.05156
204.184 2.14243
204.404 2.25907
204.637 2.39957
204.882 2.5005
205.139 2.6483
205.413 2.83829
205.703 2.95473
206.006 3.09354
206.324 3.25279
206.654 3.35028
206.996 3.47277
207.351 3.61835
207.718 3.71539
208.097 3.84825
208.491 4.0139
208.898 4.12844
209.319 4.28829
209.759 4.48853
210.215 4.62969
210.687 4.82849
211.182 5.07776
211.698 5.24292
212.233 5.45937
212.791 5.71911
213.371 5.90239
213.974 6.16324
214.605 6.48696
215.263 6.67557
215.941 6.88859
216.641 7.12243
217.36 7.25524
218.092 7.40829
218.842 7.57799
219.604 7.68039
220.378 7.8081
221.166 7.95707
221.966 8.04533
222.776 8.15031
223.597 8.26972
224.427 8.34282
225.266 8.43463
};
\addlegendentry{Omniscient Planner}
\addplot [ultra thick, TUMOrange]
table {%
156.119 7.00003
156.819 7.05934
157.531 7.22233
158.265 7.46695
159.019 7.62189
159.79 7.80638
160.581 8.01357
161.388 8.13252
162.208 8.27043
163.043 8.42387
163.889 8.50917
164.745 8.6044
165.611 8.70832
166.484 8.76579
167.364 8.82975
168.25 8.89955
169.142 8.94025
170.038 8.98926
170.94 9.04594
171.846 9.08078
172.756 9.12466
173.671 9.1769
174.583 8.98652
175.456 8.415
176.256 7.54643
176.983 6.96196
177.643 6.22665
178.224 5.38399
178.739 4.89416
179.2 4.3244
179.602 3.70142
179.955 3.35044
180.271 2.94996
};
\addlegendentry{Baseline Planner \cite{frenetix}}
\addplot [ultra thick, black, dashed]
table {%
156.119 7.00003
156.819 7.05934
157.531 7.22233
158.265 7.46695
159.019 7.62189
159.79 7.80638
160.581 8.01357
161.388 8.13252
162.208 8.27043
163.043 8.42387
163.889 8.50917
164.745 8.6044
165.611 8.70832
166.484 8.76579
167.364 8.82975
168.25 8.89955
169.142 8.94025
170.038 8.98926
170.94 9.04594
171.846 9.07409
172.755 9.09852
173.666 9.11945
174.571 8.91514
175.436 8.33782
176.228 7.47015
176.948 6.88902
177.601 6.15955
178.176 5.32464
178.685 4.83971
179.141 4.27592
179.538 3.65965
179.887 3.31253
180.199 2.91651
};
\addlegendentry{Frenetix Occlusion \cite{Moller2024}}

\node[inner sep=0pt] at (axis cs:180.199, 2.91651) {\crashIconsmall};

\addplot[draw=Grey, ultra thick, dotted] coordinates {(173,0) (173,13)};
\addlegendentry{bike becomes visible}


\legend{};

\end{axis}



\begin{axis}[
name=s_risk_plot,
at={(s_v_plot.below south)},
yshift=-0.1cm,
anchor=north,
/pgf/number format/.cd,
1000 sep={},
height=3.0cm,
width=7.5cm,
legend style={
	at={(0.5,-0.25)}, 
	anchor=north,
	legend columns=2,
	cells={anchor=center},
	draw=none,
	column sep=0.25em,
	row sep=0.1em,
},
scale only axis,
scaled ticks=false,
scaled ticks=false,
tick label style={/pgf/number format/fixed},
xlabel={Coordinate $s$ in \si{\meter}},
ylabel={Risk $R$},
x label style={at={(0.5,-0.1)},anchor=north},
xmin=157, xmax=205,
ymin=0, ymax=0.49,
ytick = {0.10, 0.20, 0.30, 0.40},
ymajorgrids=true,
]

\addplot [ultra thick, black, solid]
table {%
156.119 0
156.819 0
157.531 0
158.265 0
159.019 0
159.79 0
160.581 0.000164965
161.388 0.000164965
162.208 0.000164965
163.043 0.00145017
163.889 0.00145017
164.742 0.00145017
165.602 0.00227405
166.465 0.00227405
167.332 0.00227405
168.201 0.0912333
169.072 0.0912333
169.94 0.0912333
170.803 0.0989391
171.657 0.0989391
172.494 0.0989391
173.303 0
174.082 0
174.814 0
175.476 0
176.072 0
176.611 0
177.083 0.0361982
177.502 0.0361982
177.882 0.0361982
178.224 0.0666984
178.536 0.0666984
178.825 0.0666984
179.088 0.0980591
179.33 0.0980591
179.555 0.0980591
179.763 0.110768
179.958 0.110768
180.139 0.110768
180.306 0.138405
180.46 0.138405
180.604 0.138405
180.735 0.149689
180.856 0.149689
180.968 0.149689
181.07 0.158202
181.164 0.158202
181.251 0.158202
181.33 0.000126358
181.406 0.000126358
181.489 0.000126358
181.59 0.00211536
181.712 0.00211536
181.85 0.00211536
182.008 0.0353353
182.182 0.0353353
182.372 0.0353353
182.583 0.0962094
182.811 0.0962094
183.054 0.0962094
183.312 0.0922496
183.584 0.0922496
183.866 0.0922496
184.161 0.0763939
184.466 0.0763939
184.78 0.0763939
185.102 0.0986018
185.432 0.0986018
185.767 0.0986018
186.106 0.0794594
186.45 0.0794594
186.796 0.0794594
187.144 0.0987727
187.494 0.0987727
187.844 0.0987727
188.195 0.0781588
188.546 0.0781588
188.896 0.0781588
189.245 0.0855271
189.593 0.0855271
189.939 0.0855271
190.283 0
190.625 0
190.966 0
191.306 0
191.646 0
191.986 0
192.326 0
192.665 0
193.004 0
193.342 0
193.678 0
194.013 0
194.345 0
194.675 0
195.002 0
195.322 0
195.636 0
195.946 0
196.25 0
196.55 0
196.845 0
197.135 0
197.42 0
197.699 0
197.969 0
198.232 0
198.489 0
198.738 0
198.982 0
199.219 0
199.449 0
199.671 0
199.886 0
200.092 0
200.291 0
200.48 0
200.66 0
200.83 0
200.987 0
201.129 0
201.255 0
201.37 0
201.47 0
201.558 0
201.637 0
201.706 0.00713533
201.767 0.00713533
201.826 0.00713533
201.887 0.00063373
201.951 0.00063373
202.018 0.00063373
202.09 0.00175015
202.164 0.00175015
202.243 0.00175015
202.325 0.000170038
202.41 0.000170038
202.499 0.000170038
202.59 0.00596034
202.683 0.00596034
202.779 0.00596034
202.878 0.0749679
202.98 0.0749679
203.088 0.0749679
203.204 0.0831205
203.329 0.0831205
203.461 0.0831205
203.605 0.0908153
203.757 0.0908153
203.918 0.0908153
204.088 0.093885
204.266 0.093885
204.451 0.093885
204.643 0
204.843 nan
205.053 nan
205.278 nan
205.517 nan
205.768 nan
206.033 nan
206.311 nan
206.6 nan
206.902 nan
207.216 nan
207.54 nan
207.877 nan
208.226 nan
208.587 nan
208.966 nan
209.36 nan
209.772 nan
210.205 nan
210.659 nan
211.135 nan
211.639 nan
212.17 nan
212.726 nan
213.313 nan
213.927 nan
214.564 nan
215.227 nan
215.912 nan
216.613 nan
217.331 nan
218.065 nan
218.81 nan
219.571 nan
220.343 nan
221.127 nan
221.924 nan
222.733 nan
223.551 nan
224.381 nan
225.219 0
};
\addlegendentry{Ours $(R_\mathrm{max} = 0.10)$}

\addplot [ultra thick, TUMBlue, solid]
table {%
156.119 0
156.819 0
157.531 0
158.265 0
159.019 0
159.79 0
160.581 0.000164965
161.388 0.000164965
162.208 0.000164965
163.043 0.0162801
163.889 0.0162801
164.745 0.0162801
165.611 0.00840141
166.484 0.00840141
167.364 0.00840141
168.25 0.392216
169.142 0.392216
170.038 0.392216
170.94 0
171.839 0
172.7 0
173.489 0
174.205 0
174.857 0
175.43 0
175.938 0
176.393 0
176.789 0.106675
177.139 0.106675
177.463 0.106675
177.764 0.393194
178.051 0.393194
178.337 0.393194
178.625 0.470672
178.92 0.470672
179.229 0.470672
179.562 0.449302
179.918 0.449302
180.295 0.449302
180.697 0.292011
181.12 0.292011
181.561 0.292011
182.021 0.0754827
182.496 0.0754827
182.983 0.0754827
183.483 0.215453
183.993 0.215453
184.51 0.215453
185.036 0.17126
185.567 0.17126
186.103 0.17126
186.642 0.13843
187.183 0.13843
187.725 0.13843
188.267 0.129015
188.807 0.129015
189.345 0.129015
189.88 0
190.411 0
190.938 0
191.46 0
191.977 0
192.488 0
192.99 0
193.485 0
193.972 0
194.45 0
194.921 0
195.382 0
195.831 0
196.268 0
196.696 0
197.112 0
197.518 0
197.914 0
198.297 0
198.667 0
199.025 0
199.364 0
199.686 0
199.99 0
200.272 0
200.534 0
200.774 0
200.988 0
201.178 0
201.349 0
201.499 0.00344271
201.632 0.00344271
201.756 0.00344271
201.875 0.000108049
201.992 0.000108049
202.108 0.000108049
202.224 0.00583415
202.34 0.00583415
202.458 0.00583415
202.577 0.0803299
202.698 0.0803299
202.825 0.0803299
202.959 0.124831
203.102 0.124831
203.254 0.124831
203.418 0.145389
203.592 0.145389
203.777 0.145389
203.975 0.151333
204.184 0.151333
204.404 0.151333
204.637 0
204.882 nan
205.139 nan
205.413 nan
205.703 nan
206.006 nan
206.324 nan
206.654 nan
206.996 nan
207.351 nan
207.718 nan
208.097 nan
208.491 nan
208.898 nan
209.319 nan
209.759 nan
210.215 nan
210.687 nan
211.182 nan
211.698 nan
212.233 nan
212.791 nan
213.371 nan
213.974 nan
214.605 nan
215.263 nan
215.941 nan
216.641 nan
217.36 nan
218.092 nan
218.842 nan
219.604 nan
220.378 nan
221.166 nan
221.966 nan
222.776 nan
223.597 nan
224.427 nan
225.266 0
};
\addlegendentry{Omniscient}

\addplot [ultra thick, TUMOrange]
table {%
156.119 0
156.819 0
157.531 0
158.265 0
159.019 0
159.79 0
160.581 0.000164965
161.388 0.000164965
162.208 0.000164965
163.043 0.0162801
163.889 0.0162801
164.745 0.0162801
165.611 0.00840141
166.484 0.00840141
167.364 0.00840141
168.25 0.392216
169.142 0.392216
170.038 0.392216
170.94 0.4572
171.846 0.4572
172.756 0.4572
173.671 0
174.583 0
175.456 0
176.256 0.00544364
176.983 0.00544364
177.643 0.00544364
178.224 0.0649824
178.739 0.0649824
179.2 0.0649824
179.602 0.151246
179.955 0.151246
180.271 0.151246
};
\addlegendentry{Baseline Planner \cite{frenetix}}

\addplot [ultra thick, black, dashed]
table {%
156.119 nan
156.819 nan
157.531 nan
158.265 nan
159.019 nan
159.79 nan
160.581 nan
161.388 nan
162.208 nan
163.043 nan
163.889 nan
164.745 nan
165.611 nan
166.484 nan
167.364 nan
168.25 nan
169.142 nan
170.038 nan
170.94 nan
171.846 nan
172.755 nan
173.666 nan
174.571 nan
175.436 nan
176.228 nan
176.948 nan
177.601 nan
178.176 nan
178.685 nan
179.141 nan
179.538 nan
179.887 nan
180.199 nan
};
\addlegendimage{ultra thick, black, dashed}
\addlegendentry{Frenetix Occlusion \cite{Moller2024}}

\node[inner sep=0pt] at (axis cs:180.199, 0.151246) {\crashIconsmall};

\addplot[draw=Grey, ultra thick, dotted] coordinates {(173,0) (173,0.6)};
\addlegendentry{bike becomes visible}

\addplot [only marks, mark=x, mark size=0pt, mark options={color=white, thick}] coordinates {(180.199, 0.151246)};
\addlegendentry{collision w/ bike}

\end{axis}

\node[inner sep=0pt] at (4.0, -5.35) {\crashIcon};

    \end{tikzpicture}
    \vspace{-0.4cm}
        \caption{Velocity and risk profiles for the T-junction scenario with a real cyclist emerging from an occlusion behind the truck, comparing different planning approaches. The dotted line marks the point at which the cyclist becomes visible.}
    \label{fig:result_comparison_velocity_risk}
\end{figure}

Velocity and risk profiles across the scenario progression are analyzed, as shown in Fig.~\ref{fig:result_comparison_velocity_risk}. 
For the baseline and \cite{Moller2024} planners, deceleration occurs only after the object becomes visible (dotted line), leading to a situation where the EV cannot brake in time due to its high speed. In contrast, our proposed method behaves similarly to the omniscient planner by decelerating proactively before the object is visible. 

In the later stages of the scenario (see \Cref{fig:qualitative_40}), our approach keeps a lower speed in $s=\SIrange{185}{200}{\meter}$ because potential risks are identified behind the truck and the parked vehicle, where a pedestrian or cyclist could emerge. This behavior is not observed in the omniscient planner, which has complete knowledge that no such object exists in these occluded areas.

The results demonstrate that the inclusion of PAs and dynamically generated spawn points effectively addresses potential risks from occluded areas $\mathcal{A}_\mathrm{o}^\mathcal{L}$.

\subsection{Impact of Occlusion Tracking on PA Generation}

\begin{figure}[!ht]
    \centering
    \hspace{1mm}
    \begin{tikzpicture}[font=\scriptsize]
        \node[inner sep=0pt] at (0.6,0) {\includegraphics[height=2.5mm]{input/ego.png}};
        \node[align=left, anchor=west] at (0.9,0) {ego \\ vehicle};
    
        \node[inner sep=0pt] at (2.3,0) {\includegraphics[height=2.5mm]{input/stat_obstacle.png}};
        \node[align=left, anchor=west] at (2.6,0) {static \\ obstacle};
    
        \draw[thick, black, line width=1pt] (3.8,0) -- (4.1,0);
        \node[align=left, anchor=west] at (4.2, 0) {trajectory};

        \node[inner sep=0pt] at (5.8,0) {\visibleArea};
        \node[align=left, anchor=west] at (6.0,0) {$\mathcal{A}_\mathrm{v}^{\mathcal{L}}$};

        \node[inner sep=0pt] at (7.0,0) {\occludedArea};
        \node[align=left, anchor=west] at (7.2,0) {$\mathcal{A}_\mathrm{o}^\mathcal{L}$};
        
    \end{tikzpicture}
    \fbox{\includegraphics[width=0.4\textwidth, trim={9.5cm 5.5cm 9.5cm 7cm},clip]{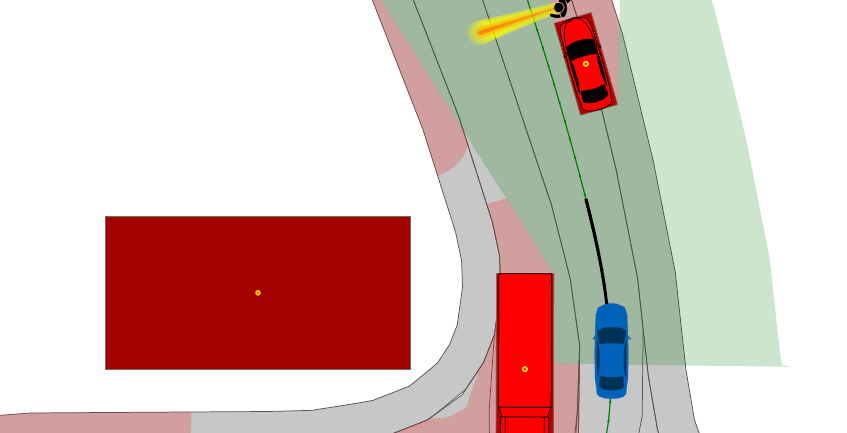}}
    \caption{Visualization of the occlusion tracking impact. Previously visible areas, determined to be obstacle-free, are excluded from the occluded area $\mathcal{A}_\mathrm{o}^\mathcal{L}$, preventing unnecessary PA generation.}
    \label{fig:result_occlusion_tracking}
\end{figure}

A qualitative analysis of the occlusion tracking impact is shown in \Cref{fig:result_occlusion_tracking}, where areas that were previously visible and determined to be free of obstacles are excluded from $\mathcal{A}_\mathrm{o}^\mathcal{L}$. These areas, depicted without the red occlusion overlay, pose no risk and do not generate PAs.

\begin{figure}[!t]
    \centering
    %
    \begin{tikzpicture}[font=\footnotesize]

\begin{axis}[
name=occlusion_tracking,
/pgf/number format/.cd,
1000 sep={},
height=3.2cm,
width=6.8cm,
legend style={
	at={(0.5,-0.25)}, 
	anchor=north,
	legend columns=2,
	cells={anchor=center},
	draw=none,
	column sep=0.25em,
	row sep=0.1em,
},
scale only axis,
scaled ticks=false,
scaled ticks=false,
xlabel={Timestep $t$},
ylabel={Occluded Area $\mathcal{A}_\mathrm{o}^\mathcal{L}$ in \si{\square\meter}},
x label style={at={(0.5,-0.1)},anchor=north},
xmin=0, xmax=90,
ymin=0, ymax=1000,
ymajorgrids=true
]
\addplot [ultra thick, black]
table {%
0	649.728678078184
3	653.740092894113
6	656.786029158572
9	657.644705491454
12	656.182904832642
15	640.199853365252
18	606.844935058015
21	507.083063072024
24	329.942585752365
27	207.542354852664
30	208.588893132175
33	192.264739301263
36	196.71158809385
39	139.368061099207
42	107.503398462856
45	114.198425706583
48	121.78804405712
51	129.695135352414
54	122.820781060714
57	117.501337023943
60	125.394021880465
63	137.416355747041
66	145.341337122505
69	149.9639701102
72	159.149250597928
75	167.651874572999
78	175.722094354939
81	181.071180284205
84	178.834200541495
87	176.766098312682
90	175.095943838904
93	169.030225666361

};
\addlegendentry{w/ tracking}
\addplot [ultra thick, TUMBlue]
table {%
0	649.728678078184
3	654.279434229925
6	656.885864593984
9	657.741535166943
12	656.279543261605
15	640.683916726413
18	607.336745507784
21	508.14606415318
24	337.651833090024
27	254.715059935474
30	595.657941372974
33	525.766039475322
36	742.16617127801
39	768.173552799654
42	654.567989940368
45	682.851085099549
48	679.148476311707
51	677.547639409529
54	634.532287568299
57	646.933814860174
60	931.770773215701
63	937.049113471885
66	877.512129142661
69	743.761510682639
72	649.187510600997
75	589.187993365701
78	547.30989404641
81	512.776082005994
84	480.527865222663
87	454.776243100134
90	422.893342943109
93	349.829512565516

};
\addlegendentry{w/o tracking}

\end{axis}

\end{tikzpicture}%

    \vspace{-0.3cm}
    \caption{Comparison of the total occluded area $\mathcal{A}_\mathrm{o}^\mathcal{L}$ over time with and without occlusion tracking for the T-junction scenario. Unlike the baseline method (w/o tracking), which assumes that large areas could suddenly be occupied, our method (w/ tracking) concludes that only a part of the area is potentially occupied.}
    \label{fig:results_occlusion_tracking}
\end{figure}
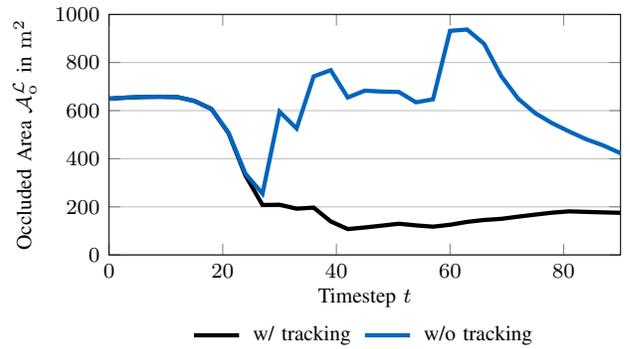

To quantitatively assess the effect of tracking, \Cref{fig:results_occlusion_tracking} compares the evolution of $\mathcal{A}_\mathrm{o}^\mathcal{L}$ in the T-junction scenario with and without occlusion tracking. 
The tracked case results in a substantially smaller occluded area throughout the scenario, as previously visible areas are excluded from further consideration. Compared to the approach in \cite{Moller2024}, our method focuses more accurately on relevant occluded areas.

Finally, the aggregated effect of occlusion tracking is evaluated across seven selected CommonRoad~\cite{commonroad} scenarios, summarized in \Cref{tab:Occluded_Area}.
\begin{table}[!ht]
\centering
\caption{Reduction of $\mathcal{A}_\mathrm{o}^\mathcal{L}$ with Occlusion Tracking (OT)}
\resizebox{0.78\linewidth}{!}{%
\begin{tabularx}{0.89\linewidth}{c c c c}
    \toprule
    Scenario & $\sum \mathcal{A}_\mathrm{o}^\mathcal{L}$ (OT) & $\sum \mathcal{A}_\mathrm{o}^\mathcal{L}$ &  Reduction\\
    \midrule
    1 & \SI{7961.7}{\square\meter} & \SI{11953.8}{\square\meter} & \SI{33.40}{\percent} \\
    2 & \SI{2780.8}{\square\meter} & \SI{4759.1}{\square\meter} & \SI{41.57}{\percent} \\
    3 & \SI{2661.7}{\square\meter} & \SI{2744.5}{\square\meter} & \SI{3.02}{\percent} \\
    4 & \SI{1115.1}{\square\meter} & \SI{1861.7}{\square\meter} & \SI{40.10}{\percent} \\
    5 & \SI{1326.9}{\square\meter} & \SI{2880.5}{\square\meter} & \SI{53.94}{\percent} \\
    6 & \SI{987.7}{\square\meter} & \SI{1659.7}{\square\meter} & \SI{40.49}{\percent} \\
    7 & \SI{2308.7}{\square\meter} & \SI{3446.0}{\square\meter} & \SI{33.00}{\percent} \\
    \bottomrule
\end{tabularx}
\label{tab:Occluded_Area}
}
\end{table}
The cumulative occluded area $\mathcal{A}_\mathrm{o}^\mathcal{L}$ is computed over the scenario duration, both with and without tracking. Particularly in scenarios involving intersections, occlusion tracking (OT) achieves an $\mathcal{A}_\mathrm{o}^\mathcal{L}$ reduction of up to \SI{54}{\percent}. This demonstrates that tracking enables the planner to focus on critical areas.

\section{Discussion}
\label{sec:discussion}

The simulation results highlight how our proposed method enables safer navigation in urban environments with occlusions. By combining occlusion tracking with PA generation, our approach addresses occluded areas while avoiding overly conservative behavior. A key advancement over \cite{Moller2024} lies in the targeted placement of PAs. Instead of considering all occluded areas, PAs are generated only in areas deemed critical to the EV’s planned trajectory. This focused approach reduces unnecessary complexity while allowing a more detailed safety assessment for high-risk areas. By incorporating multiple speed and acceleration profiles, the algorithm further ensures that a wide range of potential agent behaviors is accounted for.

The evaluation scenarios were carefully selected to replicate real-world accident-prone situations. However, additional tests across more diverse environments remain necessary to validate generalizability. Moreover, the selection of criticality thresholds remains a challenging aspect. A key limitation of the current implementation lies in its computational efficiency. It is not yet optimized for real-time deployment on a real-world vehicle. Specifically, the sequential trajectory evaluation during safety assessment can result in significant computational overhead, particularly in complex scenarios with numerous occlusions and high PA densities.

\section{Conclusion \& Outlook}
\label{sec:conclusion}
This work presents an integrated occlusion-aware planning algorithm that advances the state of the art in AV motion planning. By combining sequential occlusion tracking with PA generation, our approach addresses the challenge of navigating occluded environments. The proposed methodology improves upon previous methods such as \cite{Moller2024} by incorporating multiple speed and behavior profiles, enabling a more nuanced evaluation of potential hazards. Simulations demonstrate the methods’s ability to react proactively to occluded areas while avoiding overly conservative behavior.

Despite its promising results, the current implementation serves as a proof of concept and requires further refinement for deployment in real-world vehicles. First, the Python-based implementation must be re-engineered into a more efficient language, such as C++, to achieve real-time performance. Second, the sequential safety assessment process should be integrated into the motion planning pipeline itself. By discarding infeasible trajectories during the trajectory generation phase, computational costs can be significantly reduced, and responsiveness improved.

Future work will also focus on validating the method in real-world scenarios using a real vehicle. Furthermore, expanding the framework to include richer traffic rules and environmental constraints could improve performance, particularly in complex urban settings. These enhancements, combined with broader testing in diverse scenarios, will ensure the method's robustness and practical applicability.

\bibliographystyle{IEEEtran}
\bibliography{literatur}
\end{document}